\documentclass[runningheads]{llncs}

\usepackage{eccv}

\usepackage{eccvabbrv}

\usepackage{graphicx}
\usepackage{booktabs}
\usepackage{comment}
\usepackage{amsmath,amssymb} % define this before the line numbering.
\usepackage{epsfig}
\usepackage{amsmath}
\usepackage{amssymb}
\usepackage{mathrsfs}
\usepackage{makecell}
\usepackage{multirow}
\usepackage{algorithmicx}  
\usepackage{algorithm}
\usepackage{algpseudocode}  
\usepackage{enumitem}
\usepackage{bm}
\usepackage{colortbl}
\usepackage{arydshln}
\usepackage{pifont}
\usepackage{wrapfig}

\usepackage[accsupp]{axessibility}  % Improves PDF readability for those with disabilities.

\usepackage[pagebackref,breaklinks,colorlinks,citecolor=eccvblue]{hyperref}

\usepackage{orcidlink}

\renewcommand{\paragraph}[1]{\noindent\textbf{#1}}
\definecolor{mycolor}{HTML}{E7EFFA}
\begin{document}

\title{OV-Uni3DETR: Towards Unified Open-Vocabulary 3D Object Detection via Cycle-Modality Propagation}
%\title{OV-Uni3DETR: Unified Open-Vocabulary 3D Object Detection via Cycle-Modality Propagation} 

\titlerunning{OV-Uni3DETR}

\makeatletter
\def\thanks#1{\protected@xdef\@thanks{\@thanks
        \protect\footnotetext{#1}}}
\makeatother

\author{Zhenyu Wang\inst{1} \quad
Yali Li\inst{1*} \quad
Taichi Liu\inst{2} \\
Hengshuang Zhao\inst{3*} \quad
Shengjin Wang\inst{1} \thanks{$^*$Corresponding author.}}

\authorrunning{Z. Wang et al.}

\institute{$^1$Department of Electronic Engineering, Tsinghua University \\ Beijing National Research Center for Information Science and Technology (BNRist) \\ $^2$Rutgers University \quad $^3$The University of Hong Kong \\
\email{wangzy20@mails.tsinghua.edu.cn} \quad
\email{\{liyali13, wgsgj\}@tsinghua.edu.cn} \\
\email{tl821@rutgers.edu} \quad
\email{hszhao@cs.hku.hk}
}

\maketitle

\begin{abstract}
In the current state of 3D object detection research, the severe scarcity of annotated 3D data, substantial disparities across different data modalities, and the absence of a unified architecture, have impeded the progress towards the goal of universality. In this paper, we propose \textbf{OV-Uni3DETR}, a unified open-vocabulary 3D detector via cycle-modality propagation. Compared with existing 3D detectors, OV-Uni3DETR offers distinct advantages: 1) Open-vocabulary 3D detection: During training, it leverages various accessible data, especially extensive 2D detection images, to boost training diversity. During inference, it can detect both seen and unseen classes.  2) Modality unifying: It seamlessly accommodates input data from any given modality, effectively addressing scenarios involving disparate modalities or missing sensor information, thereby supporting test-time modality switching. 3) Scene unifying: It provides a unified multi-modal model architecture for diverse scenes collected by distinct sensors. Specifically, we propose the cycle-modality propagation, aimed at propagating knowledge bridging 2D and 3D modalities, to support the aforementioned functionalities. 2D semantic knowledge from large-vocabulary learning guides novel class discovery in the 3D domain, and 3D geometric knowledge provides localization supervision for 2D detection images. OV-Uni3DETR achieves the state-of-the-art performance on various scenarios, surpassing existing methods by more than 6\% on average. Its performance using only RGB images is on par with or even surpasses that of previous point cloud based methods.  Code is available at \url{https://github.com/zhenyuw16/Uni3DETR}.
\end{abstract}
%on par with

\section{Introduction}

\begin{figure}[t]
\centering
\setlength{\abovecaptionskip}{0pt}
\setlength{\belowcaptionskip}{0pt}
\begin{subfigure}{0.48\textwidth}
\centering
\includegraphics[width=\textwidth]{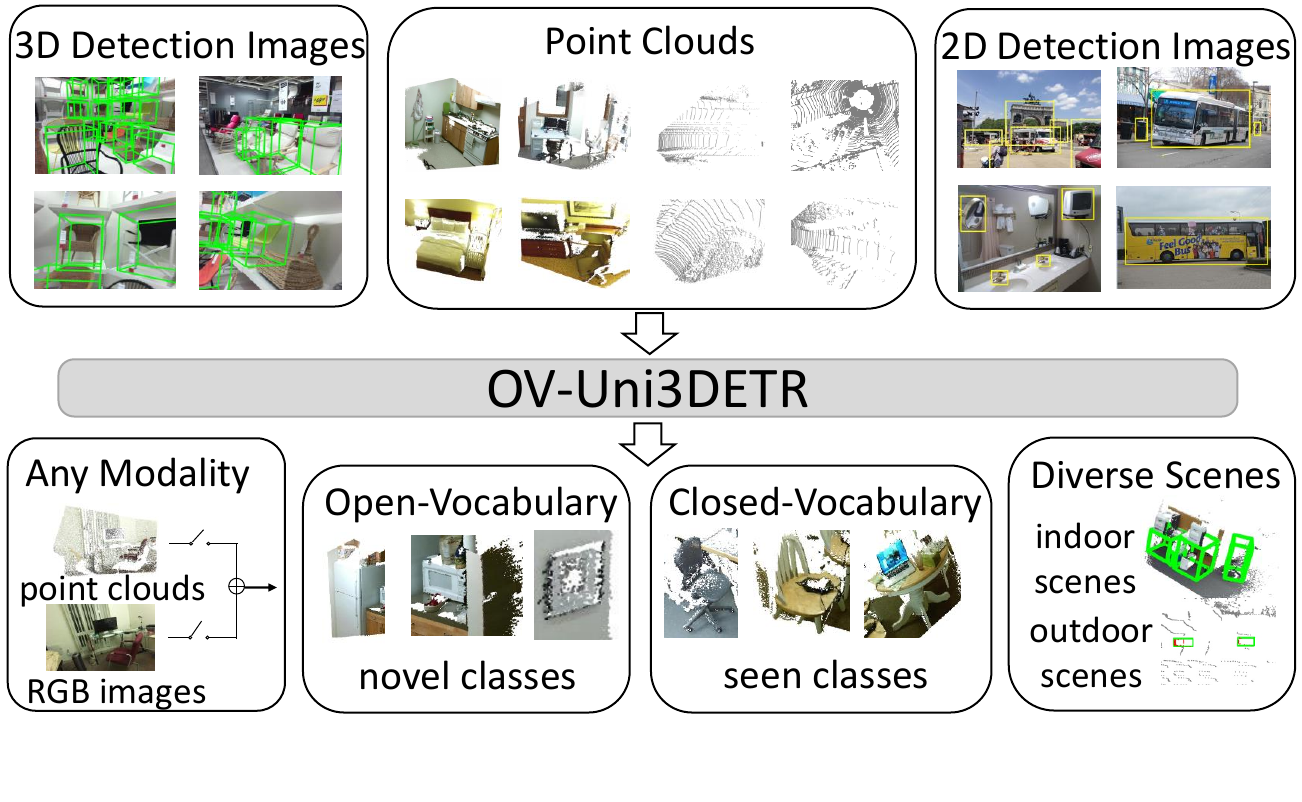}
\caption{What can OV-Uni3DETR do?}
\label{fig:m1}
\end{subfigure}
\hfill
\begin{subfigure}{0.5\textwidth}
\centering
\includegraphics[width=\textwidth]{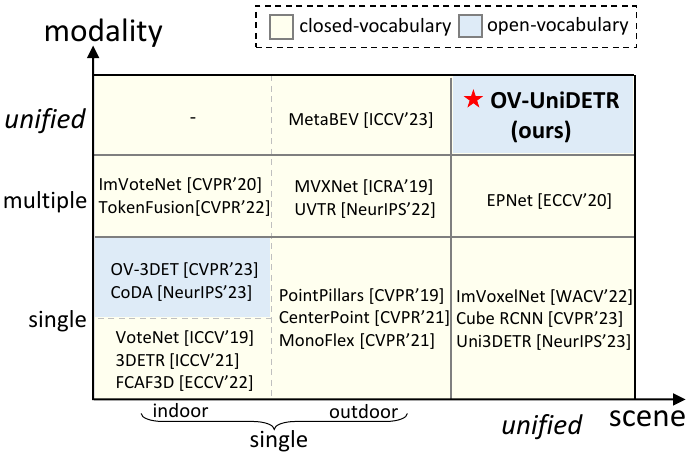}
\caption{How does OV-Uni3DETR distinguish?}
\label{fig:m2}
\end{subfigure}
\caption{\textbf{Illustration for OV-Uni3DETR.} (a): It utilizes various available data for training, including 3D point clouds, 3D detection images (with 3D box annotated and aligning with point clouds) and 2D detection images (only 2D box annotated). For inference, it can predict 3D boxes using any modality data, for both open-vocabulary and closed-vocabulary, both indoor and outdoor 3D detection. (b): Compared with existing 3D detectors, OV-Uni3DETR achieves modality unifying (modality-switchable during inference), scene unifying, and open-vocabulary learning simultaneously. }
\label{fig:overview}
\end{figure}

3D object detection aims to predict the oriented 3D bounding boxes and the semantic category tags for the real scenes given point clouds or RGB images. Recently, 2D object detection for universality has developed rapidly \cite{zhou2022detecting, wang2023detecting, li2022grounded, liu2023grounding}. However, related research on 3D detection remains significantly lagging behind that of 2D. Currently, most of 3D object detection methods are still restricted by data fully annotated for specific input modalities, and can only recognize categories that appear during training for either indoor or outdoor scenes.

% These methods leverage various forms of available data for training, enhancing the detector's universality, and enabling it to detect any category in any scene.
%3D universal detection remains significantly lagging behind 2D

The problem of 3D universal object detection is challenging because of the following reasons. First, existing 3D detectors work in a closed-vocabulary manner, thus can only detect seen classes. \emph{Open-vocabulary 3D object detection} is urgently demanded to recognize and localize object instances with novel categories that are not acquired during training. However, the generalization ability in localizing novel objects is restricted by existing 3D detection datasets  \cite{song2015sun, dai2017scannet, geiger2013vision}, where both size and categories are limited compared to those 2D ones \cite{lin2014microsoft, shao2019objects365, kuznetsova2020open, gupta2019lvis}. Additionally, the lack of pre-trained image-text models \cite{radford2021learning, jia2021scaling, zhai2022lit} in the 3D domain further exacerbates the challenges associated with open-vocabulary 3D detection. Second, a \emph{unified architecture for multi-modal 3D detection} is absent. Existing 3D detectors are predominantly designed for specific input modalities (either point clouds, RGB images, or both) and scenes (either indoor or outdoor ones). For open-vocabulary 3D detection, especially under the constraints posed by limited data, the lack of a unified multi-modal structure prevents the effective utilization of data from various modalities and sources. Consequently, the detector cannot generalize to novel objects effectively. Besides, without a unified architecture, 3D detectors struggle to adapt to different input modalities or handle situations where sensor modalities are missing. A unified solution to address the open-vocabulary challenge in a multi-modal context, therefore, is a crucial step for current research in 3D detection.

We propose \textbf{OV-Uni3DETR}, a unified multi-modal 3D detector for open-vocabulary 3D object detection, as is in Fig. \ref{fig:m1}. During training, it is endowed with the capability to leverage multi-modal and multi-source data, including point clouds, 3D detection images with precise 3D box annotations and aligning with point clouds, and 2D detection images with only 2D box annotations. A critical enhancement is the integration of 2D detection images, which is particularly advantageous for open-vocabulary 3D detection due to the significantly greater number of annotated classes. With these multi-modal data, we further adopt a switched-modality training scheme. Benefiting from the above multi-modal learning manner, OV-Uni3DETR accommodates data from any modality during inference, thus achieving the function of test-time modality switching. It excels in detecting both base classes and novel classes. The unified structure further equips OV-Uni3DETR with the ability to detect in both indoor and outdoor scenes. As can be seen in Fig. \ref{fig:m2}, OV-Uni3DETR achieves scene and modality unifying and possesses the open-vocabulary capability, thus greatly advancing the universality of 3D detectors across categories, scenes, and modalities.

With multi-modal learning, two challenging problems should be further addressed. The first is about \emph{how to generalize the detector to novel classes}, and the second is about \emph{how to learn from the extensive 2D detection images without 3D box annotations}. We propose the approach of cycle-modality propagation - knowledge propagation between 2D and 3D modalities to address the two challenges. For 2D to 3D propagation, we extract 2D bounding boxes with a 2D open-vocabulary detector, and project them into the point cloud space to approximate 3D boxes. In this way, the abundant semantic knowledge from the 2D detector can be propagated to the 3D domain to assist novel box discovery. For 3D to 2D, we leverage the geometric knowledge from a class-agnostic 3D detector to localize objects within 2D detection images, and assign category tags by Hungarian matching. Such geometric knowledge can compensate for the absence of 3D supervision information in 2D detection images.

Our main contributions can be summarized as follows:

\begin{itemize}[topsep=0pt, parsep=0pt, itemsep=0pt, partopsep=0pt, leftmargin=15pt]
    \item We propose OV-Uni3DETR, a unified open-vocabulary 3D detector with multi-modal learning. It excels in detecting objects of any class across various modalities and diverse scenes, thus greatly advancing existing research towards the goal of universal 3D object detection.
    \item We present a unified multi-modal architecture for both indoor and outdoor scenes. By eliminating modality inconsistencies and switched-modality training, it becomes test-time modality-switchable: utilizing any modality data.
    \item We propose the concept of a knowledge propagation cycle between 2D and 3D modalities. By leveraging 2D large-vocabulary semantic knowledge and precise 3D geometric knowledge, the training diversity can be guaranteed.  % for open-vocabulary detection.
\end{itemize}
% and the future of 3D foundation models

% Extensive experiments demonstrate the strong ability of OV-Uni3DETR. It achieves state-of-the-art performance on various tasks of 3D detection. With universality in modality, scene and category, we believe OV-Uni3DETR can become a significant step towards the future of 3D foundation models.
%In the open-vocabulary setting, it surpasses previous methods by more than 7\% on SUN RGB-D \cite{song2015sun} and 8\% on ScanNet \cite{dai2017scannet}. For traditional close-vocabulary experiments, OV-Uni3DETR is also more than 3\% higher than previous methods. It is the first model that is able to perform open-vocabulary detection on outdoor 3D datasets. 

\section{Related Work}

\paragraph{3D Object Detection} aims to predict category tags and oriented 3D bounding boxes for the scene. Some works take point cloud data as input. Because of the significant distinction in point cloud data, these models are usually separated into indoor  \cite{qi2019deep, zhang2020h3dnet, liu2021group, rukhovich2022fcaf3d, wang2022cagroup3d} and outdoor \cite{yan2018second, shi2020points, shi2020pv, sheng2021improving, yin2021center} scenarios. A unified structure \cite{wang2023uni3detr} for both indoor and outdoor point clouds is presented recently. RGB-based 3D detectors \cite{rukhovich2022imvoxelnet, brazil2023omni3d, liu2020smoke, wang2021fcos3d, zhang2021objects, li2022bevformer} utilize only RGB images for 3D bounding box prediction. Constrained by the limited spatial information in RGB images, the performance of these detectors significantly lags behind their counterparts using point clouds. In comparison, multi-modal 3D detectors \cite{sindagi2019mvx, qi2020imvotenet, liang2022bevfusion, wang2022multimodal, Ge_2023_ICCV} utilize both point cloud data and RGB images. With multi-modal data, these detectors make better performance. However, these models can only work in the closed-vocabulary setting, thus are restricted from the limited scale of 3D detection data. A unified structure for all modalities and scenes is also absent currently.

\paragraph{Open-Vocabulary Object Detection} aims to recognize and localize novel classes that are not annotated in datasets. Benefiting from large-scale image-text pre-training models \cite{radford2021learning, jia2021scaling, zhai2022lit}, 2D open-vocabulary detection researches have been forwarded significantly. \cite{zareian2021open, gu2021open, du2022learning, ma2022open, minderer2022simple, wang2023detecting} adopt such pre-trained parameters and detect novel classes with the help of text features. \cite{zhou2022detecting, bangalath2022bridging} involve image-level supervision to help expand the vocabulary of the detectors. \cite{li2022grounded, zhang2022glipv2, liu2023grounding} unify object detection and visual grounding for pre-training and adopt text queries for detection. These methods focus on 2D detection. In the 3D field, restricted by the limited scare of 3D data, there are still no pre-trained point-text models, which makes 3D open-vocabulary detection a challenging problem.

\paragraph{Open-Vocabulary 3D Object Detection} targets at novel class recognition and localization for 3D bounding boxes. Some works \cite{zhang2022pointclip, zeng2023clip2, ding2023pla, xue2023ulip} borrow ideas from 2D image-text pre-training and utilize text embeddings for 3D novel classes. These works are mainly about classification and semantic segmentation, and cannot be adopted in 3D detection. Recently, \cite{cen2021open, lu2022open} have forwarded open-set 3D detection, \cite{lu2023open} conducts open-vocabulary 3D detection with the help of a pre-trained 2D detector, and \cite{cao2023coda} recognizes novel classes by novel object discovery and cross-modal alignment. However, these models only utilize point cloud data for inference, and can only work in indoor scenes. In comparison, we design a unified open-vocabulary detector for multi-modal data and different scenes.

\section{OV-Uni3DETR}

\begin{figure}[t]
\centering
\includegraphics[width=\textwidth]{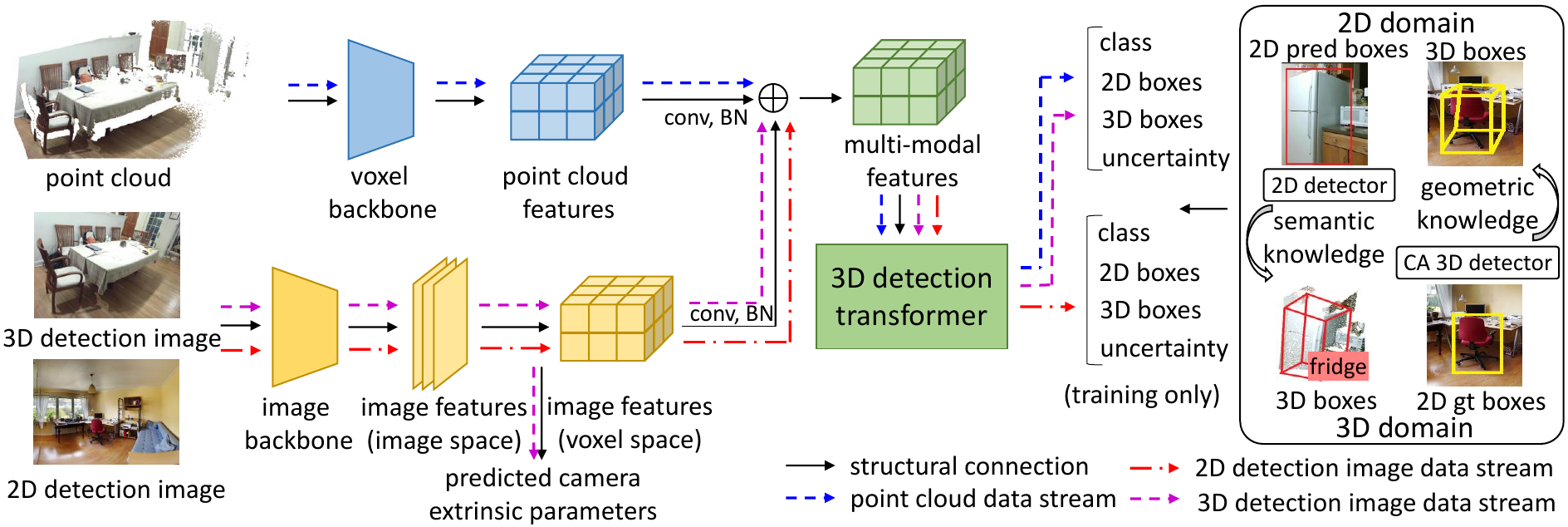}
\caption{\textbf{Overview of OV-Uni3DETR.} We extract features for point clouds and images. After converted into the same voxel space, they are added for the multi-modal features. The 3D detection transformer is finally utilized for class and box prediction. We perform semantic knowledge propagation from 2D to 3D for novel class discovery. To use 2D detection images, we predict the camera extrinsic parameters and propagate geometric knowledge from 3D to 2D through a class-agnostic (CA) 3D detector. }
\label{fig:frame}
\end{figure}

We present the overview of OV-Uni3DETR in Fig. \ref{fig:frame}. It takes both point clouds and RGB images during training. Once trained, it can be switchable and use either of them for inference. We propagate semantic knowledge from 2D to 3D for novel classes to fulfill open-vocabulary 3D detection learning. Extra 2D detection images are further introduced to promote novel class recognition, with a class-agnostic 3D detector propagating geometric knowledge from 3D to 2D.

\subsection{Multi-Modal Learning}

Our multi-modal architecture takes point cloud $X_P$ and images $X_I$ for training, and can handle situations when missing sensor modalities exists for inference, \emph{i.e.}, test-time modality-switchable. We first extract 3D point cloud features $F_P \in \mathbb{R}^{C \times X \times Y \times Z}$ with the voxel-based backbone, and 2D image features $F_I \in \mathbb{R}^{C \times H \times W}$ with the image backbone, where $(H, W)$ denotes the size of extracted image features. $F_I$ is then projected into the 3D voxel space for the image features in the voxel space $F_I' \in \mathbb{R}^{C \times X \times Y \times Z}$ through the camera parameters. Specifically, denote the camera intrinsic matrix as $K$ and the extrinsic matrix as $R_t$, then the corresponding positions in the 2D image can be obtained by projecting 3D positions in the voxel space through $KR_t$.

Compared with 2D images, 3D point cloud data usually consist of more spatially rich information, which is crucial for 3D detection. As a result, the trained multi-modal detector is easy to strongly rely on point cloud features for detection, while simply treating image features as auxiliary information. Under this manner, if point cloud data are absent at the inference time, the performance will deteriorate seriously. To avoid this problem, we first utilize a 3D convolution layer with batch normalization. The multi-modal features are thus obtained by: $F_M = {\rm BN}({\rm conv}(F_P)) + {\rm BN}({\rm conv}(F_I')$). With the 3D convolution and BN layers, different modality features are regularized. This prevents feature discrepancy and the suppression of image features. We then adopt the switched-modality training scheme to further avoid image features collapsing. Specifically, the 3D detection transformer randomly receives features from aforementioned modalities - $F_M$, $F_P$, $F_I'$, with pre-determined probabilities, which makes it possible for the model to detect with only single-modal data. By random switching, the model accepts image-only features during training. This also prevents the image features from being ignored. We directly use the 3D transformer in \cite{wang2023uni3detr} for detection.

%r to project the point cloud features and the image features into the same feature space

The multi-modal architecture finally predicts the category tags, the 4-dim 2D boxes and 7-dim 3D boxes for both 2D and 3D object detection. The L1 loss and decoupled IoU loss \cite{wang2023uni3detr} are utilized for 3D box regression, and the L1 loss and GIoU loss \cite{rezatofighi2019generalized} are utilized for 2D box regression. In the open-vocabulary setting, novel class samples exist, which increases the difficulty of training samples. We therefore introduce an uncertainty prediction $\mu$ following \cite{lu2021geometry, brazil2023omni3d}, and utilize it to weight the L1 regression loss. The loss for object detection learning is as:
\begin{equation}
    L = L_{cls}  + \sqrt{2} \cdot {\rm exp}(-\mu) \cdot (L_{1}^{3D} + L_{1}^{2D}) + L_{IoU}^{3D} + L_{IoU}^{2D} + \mu
\end{equation}

For some 3D scenes, there may exist multi-view images, rather than the single monocular one. We extract image features for each of them and project to the voxel space using their own projection matrices. The multiple image features in the voxel space are summed for the multi-modal features. 

\subsection{Knowledge Propagation: 2D $\rightarrow$ 3D}

We perform open-vocabulary 3D detection based on multi-modal learning introduced before. The core issue of open-vocabulary learning is recognizing novel classes that are not human-annotated during training. Due to the difficulty of acquiring point cloud data, the pre-trained vision-language models \cite{radford2021learning, jia2021scaling, zhai2022lit} do not exist in the point cloud field. The modality difference between point cloud data and RGB images restricts the performance of these models in 3D detection. We propose to leverage semantic knowledge from a pre-trained 2D open-vocabulary detector, and generate the corresponding 3D bounding boxes for novel classes. The generated 3D boxes will supplement 3D ground-truth labels with limited classes available at the training time.

Specifically, we first generate 2D bounding boxes or instance masks with a 2D open-vocabulary detector. Considering that the available data and annotations are much more abundant in the 2D field, these generated 2D boxes can achieve higher localization accuracy and cover a significantly wider range of categories. We then project these 2D boxes to the 3D space through $(KR_t)^{-1}$ to obtain the corresponding 3D boxes. The specific operation is that we project 3D points into the 2D space using $KR_t$, finding points within the 2D boxes, then clustering these points inside the 2D boxes for eliminating outliers to obtain the corresponding 3D boxes. Benefiting from the pre-trained 2D detector, novel objects that are not annotated can be available in the generated 3D box set. In this way, the rich semantic knowledge deriving from large-vocabulary 2D detection learning can be propagated from the 2D domain to the generated 3D boxes, thus greatly boosting 3D open-vocabulary detection. For multi-view images, 3D boxes are generated separately and ensembled together for the final usage. 

During inference, when both point clouds and images are available, we can extract 3D boxes in the similar way. These generated 3D boxes can also be viewed as a form of 3D open-vocabulary detection results. We add these 3D boxes to the predictions of the multi-modal 3D transformer to supplement the potentially missing objects, and filter the overlapped bounding boxes through 3D NMS. The confidence scores assigned by the pre-trained 2D detector are systematically divided by a predetermined constant, and then reinterpreted as the confidence scores of the corresponding 3D boxes.

\subsection{Knowledge Propagation: 3D $\rightarrow$ 2D}

\begin{figure}[t]
\centering
\includegraphics[width=\columnwidth]{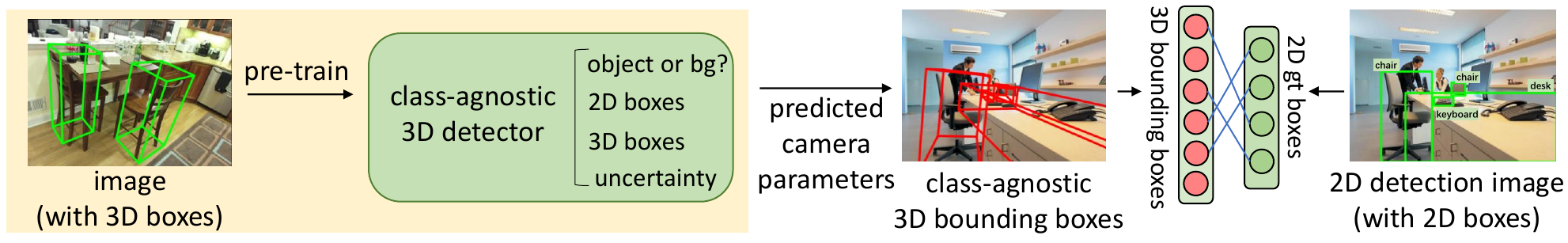}
\caption{\textbf{Illustration of knowledge propagation from 3D to 2D.} 3D detection images first train a class-agnostic 3D detector, which is then used to generate class-agnostic 3D bounding boxes with the predicted camera parameters for the 2D detection images. Hungarian matching is finally conducted between 2D boxes and 3D ones for the class-specific 3D bounding boxes.}
\label{fig:2ddetection}
\end{figure}

Compared with 3D object detection data with 3D bounding boxes annotated, 2D detection images with just 2D bounding boxes annotated are much more abundant, with significantly more scenes, objects and categories included. Motivated by this, we introduce these 2D detection images for training to boost open-vocabulary 3D detection, especially in novel class recognition. However, 3D bounding boxes are not annotated for these 2D detection images. Learning totally without 3D box annotations will have a negligible impact on 3D box prediction training, making 2D detection images hard to be fully utilized. Besides, the camera parameters are absent, making it infeasible to transform image features to the voxel space. We propose to propagate geometric knowledge from 3D to 2D through a class-agnostic 3D detector. The 3D bounding boxes and camera parameters will be predicted for these 2D detection images.

\paragraph{Camera parameter prediction.} The camera parameters mainly include the camera intrinsic matrix $K$ and the extrinsic matrix $R_t$. The intrinsic matrix $K$ is mainly about the the camera’s focal lengths $(f_x, f_y)$ and the principal point $(p_x, p_y)$. Denote the image resolution as $(h, w)$, we simply treat $f_x = f_y = h$, $p_x = \frac{1}{2}w$, $p_y = \frac{1}{2}h$. The intrinsic matrix is thus estimated as: $K = \begin{bmatrix}
h& 0 & \frac{1}{2}w \\
0& h & \frac{1}{2}h \\
0 & 0 & 1
\end{bmatrix}$.

The camera extrinsic matrix $R_t \in \mathbb{R}^{3 \times 4}$ denotes the transformation from the 3D world coordinates to 3D camera coordinates. It can be written in $R_t = [R|T]$, where $R \in \mathbb{R}^{3 \times 3}$ is the rotation matrix and $T \in \mathbb{R}^{3 \times 1}$ is the position of the origin of the world coordinate system. The rotation matrix $R$ is about rotating the angle $\theta$ around the axis $u = (u_x, u_y, u_x)$, and $T$ is specifically expressed in $T=[t_x, t_y, t_z]^T$. In total, there are 7 parameters about the extrinsic matrix $R_t$ to be estimated. We add a 8-dim branch on the image features in the image space to predict these parameters, specifically, $[{\rm sin}\theta, {\rm cos}\theta, u_x, u_y, u_x, t_x, t_y, t_z]$. We use the L1 loss for regression to learn these parameters. This camera parameter prediction module is pre-trained on 3D detection images, then utilized to predict missing camera parameters for 2D detection images. 

\paragraph{Generating 3D bounding boxes.} For 2D detection images, 2D annotated bounding boxes $\{c_i, bb^{2D}_{i}\}_{i=1}^M$ are available for object instances, where $c_i$ and $bb^{2D}_{i}$ are the category tag and 2D bounding box of the $i$-th object, $M$ is the number of 2D boxes. We aim to generate 3D bounding boxes for these objects. 

We first pre-train a class-agnostic 3D detector on 3D detection images with only seen class objects. Here we discard the point cloud branch for the class-agnostic 3D detector. The semantic category tags are directly removed for class-agnostic learning. Then, this pre-trained detector conducts inference on 2D detection images to obtain the predicted class-agnostic boxes $\{\hat{bb}^{2D}_{i}, \hat{bb}^{3D}_{i}\}_{i=1}^N$, where $N$ is the box number, using the predicted camera parameters from the pre-trained camera parameter prediction module. Since class-agnostic classification has the generalization ability for novel classes \cite{gu2021open, saito2022learning, kim2022learning, wang2023detecting}, the predicted class-agnostic 3D boxes can include novel class objects, even only base classes participating in training. With the class-agnostic 3D detector, geometric knowledge from the 3D domain can be propagated to 2D detection images, and provide 3D localization information for them. As a result, the issue of the lack of 3D supervision information can be addressed. 
%We discard the point cloud branch of the above architecture for the class-agnostic detector. 
%Here we discard the point cloud branch of our multi-modal architecture for the class-agnostic 3D detector. 

We then need to assign category tags to the class-agnostic boxes. We formulate this problem as finding a bipartite matching between the ground-truth set and the extracted 3D box set. We propose that the optimal bipartite matching should minimize the overlaps between the ground-truth 2D boxes and extracted class-agnostic 3D boxes. We calculate the IoU between $bb^{2D}_{i}$ and predicted 2D class-agnostic boxes $\hat{bb}^{2D}_{i}$, and the bipartite matching problem is written as:
\begin{equation}
    \hat{\sigma} = \underset{\sigma}{\arg\min} \sum_i^N {\rm IoU}(bb^{2D}_{i}, \hat{bb}^{2D}_{\sigma(i)} )
\end{equation}

We utilize the Hungarian algorithm to solve it \cite{carion2020end}. After bipartite matching, the category tags can be assigned to the 3D boxes to obtain the class-specific 3D bounding boxes $\{c_i, bb^{2D}_{i}, \hat{bb}^{3D}_{\sigma(i)}\}_{i=1}^M$. They are treated as the ground-truth labels of these 2D detection images for training. With the generated class-specific 3D boxes, these 2D detection images can be used in the same way as 3D detection images. However, there exists noise in 3D boxes $\hat{bb}^{3D}_{i}$, which will hurt the 3D box prediction performance. Therefore, we design a dual-branch structure for the final output layers of the OV-Uni3DETR. It specifically refers to two distinct parameter sets of output layers with the same structure and loss. 3D detection images and point cloud data pass into one branch for classification and regression, and 2D detection images use another one. In this way, the noisy 3D boxes will not interfere with the accurate ones. At the inference time, the branch for 2D detection images will be discarded for accurate inference.

\section{Experiments}

We conduct extensive experiments under various conditions for open-vocabulary and traditional closed-vocabulary settings to demonstrate the strong detection ability of our OV-Uni3DETR in this section.

\paragraph{Datasets.} For indoor 3D detection, we adopt two challenging datasets, SUN RGB-D \cite{song2015sun} and ScanNet V2 \cite{dai2017scannet}. We mainly follow the setting of \cite{cao2023coda} for open-vocabulary experiments. SUN RGB-D contains 5,285 training and 5,050 validation scenes, with 46 classes in total and oriented 3D bounding boxes annotated. Each scene consists of one single-view RGB image. The categories with the top 10 most training samples are selected as base (seen) categories, while the rest 36 are novel classes. ScanNet V2 contains 1,201 reconstructed training scans and 312 validation scans, with 200 object categories for axis-aligned bounding boxes in total \cite{rozenberszki2022language}. We adopt the same setting as \cite{cao2023coda} for training and inference, where point clouds are generated from the single-view depth images. The top 10 classes are used for base classes, and the other 50 ones are novel classes. We also conduct experiments on the ScanNet setting from \cite{lu2023open}, which includes 20 classes as novel ones, no seen classes, for further comparison. For 2D detection images, we randomly select 6,000 images from the Objects365 \cite{shao2019objects365} dataset. We mainly use the mean average precision (mAP) under IoU thresholds of 0.25. 
%We download the officially released version from \cite{cao2023coda} for training and inference, where point clouds are generated from the single-view depth images.

For outdoor 3D detection, we mainly conduct experiments on the KITTI \cite{geiger2013vision} and nuScenes \cite{caesar2020nuscenes} dataset. For KITTI, we split its official training set into 3,712 training samples and 3,769 validation samples for training and evaluation. Each scene contains one monocular image. We treat the car and cyclist classes as base classes, and the pedestrian category for the novel one. For nuScenes, we train on the 28,130 frames of samples from the training set and evaluate on the 6,010 validation samples. Each scene contains 6 images with different directions here. The classes of "car, trailer, construction vehicle, motorcycle, bicycle" are treated as seen and the rest five are unseen ones.

\paragraph{Implementation details.} We implement OV-Uni3DETR mainly with mmdetection3D \cite{contributors2020mmdetection3d}, and train it with the AdamW \cite{loshchilov2017decoupled} optimizer. The point cloud branch is the same as \cite{wang2023uni3detr}, and we use ResNet50 \cite{he2016deep} and FPN \cite{lin2017feature} for the image feature extractor without specially mentioned. For main experiments we generate 3D bounding boxes of the 365 classes of the Objects365 dataset for training. We use a pre-trained Detic \cite{zhou2022detecting} for the 2D open-vocabulary detector. All generated labels are filtered by the 0.4 confidence threshold. 

\subsection{Open-Vocabulary 3D Object Detection}

\begin{table}[t]
\centering
\setlength{\abovecaptionskip}{0pt}
\setlength{\belowcaptionskip}{0pt}
\caption{\textbf{The performance of OV-Uni3DETR on the SUN RGB-D and ScanNet dataset for open-vocabulary 3D object detection.} P denotes the point cloud inputs and I denotes the image inputs. The experimental setting is totally the same as CoDA, and the utilized data are downloaded from CoDA officially released code.}
\resizebox{0.75\columnwidth}{!}{
\begin{tabular}{c|c|ccc|ccc}
\Xhline{1.1pt} 
\multirow{2}*{Method}& \multirow{2}*{Inputs}& \multicolumn{3}{c|}{SUN RGB-D} & \multicolumn{3}{c}{ScanNet} \\
% \cline{3-8}
 &  & AP$_{novel}$ & AP$_{base}$ & AP$_{all}$ & AP$_{novel}$ & AP$_{base}$ & AP$_{all}$ \\ 
\hline
 Det-PointCLIP \cite{zhang2022pointclip} & P &0.09 & 5.04 & 1.17 & 0.13 & 2.38 & 0.50\\
 Det-PointCLIPv2 \cite{zhu2023pointclip} & P & 0.12 & 4.82 & 1.14 & 0.13 & 1.75 & 0.40 \\
 Det-CLIP$^2$ \cite{zeng2023clip2} & P & 0.88 & 22.74 & 5.63  & 0.14 & 1.76 & 0.40 \\
 3D-CLIP \cite{radford2021learning} & P+I & 3.61 & 30.56 & 9.47 & 3.74 & 14.14 & 5.47\\
 CoDA \cite{cao2023coda} & P & 6.71 & 38.72 & 13.66 & 6.54 & 21.57 & 9.04\\
\hline
\multirow[c]{3}{*}{OV-Uni3DETR (ours)} & P  & 9.66 & 48.29 & 18.06  & 12.09 & 30.47 & 15.15 \\
 & I & 5.41 & 29.51 & 10.65 & 8.10 & 20.87 & 10.23\\
 & P+I & \textbf{12.96} & \textbf{49.25} & \textbf{20.85} & \textbf{15.21} & \textbf{31.86} & \textbf{17.99}\\
\Xhline{1.1pt} 
\end{tabular}
}
\label{tab:indooropen}
\end{table}
% \footnotetext{ The experimental setting is totally the same as CoDA, where utilized data are downloaded from CoDA officially released code.}

\begin{table*}[t]
\centering 
\setlength{\abovecaptionskip}{0pt}
\setlength{\belowcaptionskip}{0pt}
\caption{\textbf{Comparison with methods in the same setting of \cite{lu2023open} on ScanNet.} Mean represents the average value of all 20 categories. OV-Uni3DETR is evaluated with the same setting, with only point clouds utilized for both training and inference.}
\resizebox{\textwidth}{!}{
\begin{tabular}{c|c|cccccccccc}
\Xhline{1.1pt} 
Methods& \textbf{Mean} & toilet  & bed  & chair & sofa & dresser & table & cabinet & bookshelf & pillow & sink \\
\hline
OV-3DET \cite{lu2023open} & 18.02 & 57.29 & 42.26 & 27.06 & 31.50 & 8.21 & 14.17 & 2.98 & 5.56 & 23.00 & 31.60  \\
CoDA \cite{cao2023coda} & 19.32 & 68.09 & 44.04 &  28.72 &  44.57 &  3.41 & 20.23 &  5.32 &  0.03 &  27.95 &  45.26     \\
OV-Uni3DETR (ours) & \textbf{25.33} & 86.05 & 50.49 & 28.11 & 31.51 & 18.22 & 24.03 & 6.58 & 12.17 & 29.62 & 54.63\\
%\Xhline{1.1pt}
\hline
 Methods & & bathtub & refrigerator & desk & nightstand & counter & door &  curtain & box & lamp  & bag  \\
\hline
 OV-3DET & & 56.28 & 10.99 & 19.72 & 0.77 & 0.31 & 9.59 & 10.53 & 3.78 & 2.11 & 2.71   \\
CoDA & &  50.51 &  6.55 &  12.42 &  15.15 &  0.68 &  7.95 &  0.01  &  2.94 &  0.51 &  2.02 \\
OV-Uni3DETR (ours) &  & 63.73 & 14.41 & 30.47 & 2.94 & 1.00 & 1.02 & 19.90 & 12.70 & 5.58 & 13.46 \\
\Xhline{1.1pt} 
    \end{tabular}
    }
    \label{tab:cmpwith3ddet}
\end{table*}

% In this subsection, we mainly evaluate the open-vocabulary performance of our OV-Uni3DETR on the indoor and outdoor 3D detection datasets. 

\paragraph{Indoor 3D open-vocabulary detection.} We evaluate OV-Uni3DETR on the indoor SUN RGB-D dataset for the 46 class setting, and list the AP$_{25}$ metric in Tab. \ref{tab:indooropen}. OV-Uni3DETR obtains 9.66\% AP$_{25}$ for the 36 novel classes with point clouds only during inference, which surpasses CoDA \cite{cao2023coda} by 2.95\% with the same used data. This demonstrates that our method recognizes novel classes well with the knowledge propagation cycle. Meanwhile, AP$_{base}$ is even 9.57\% higher, which should be credited to the multi-modal architecture. When only RGB images are available for inference, AP$_{novel}$ is 5.41\%, which is comparable to the performance of CoDA using point clouds. This demonstrates that the RGB image features do not collapse during training. As a result, RGB-only inference achieves an equally excellent 3D detection result, which well illustrates that OV-Uni3DETR can take any modality data and can be switchable for different modalities during inference. When both point clouds and RGB images are available, the detector obtains 12.96\% AP$_{novel}$, 6.25\% higher than CoDA. 
This demonstrates that our \textbf{cycle-modality propagation effectively facilitates the integration of multi-modal knowledge and the comprehensive utilization of information from different modalities}, thus assisting in achieving superior performance for open-vocabulary 3D detection. %Therefore, OV-Uni3DETR performs significantly better for open-vocabulary detection, and can utilize any modality data. 

We then evaluate OV-Uni3DETR on ScanNet and list the AP$_{25}$ metric in Tab. \ref{tab:indooropen}. We utilize the same setting as CoDA, where one single-view image corresponds to one point cloud scene. Meanwhile, since many images share high similarity in content in this setting, with only slight difference in perspective, we do not use extra 2D detection images here. We observe that the superiority of our method becomes more remarkable on the ScanNet dataset. We achieve 5.55\% higher AP$_{novel}$ than CoDA using only point cloud data, and the base category AP$_{25}$ is even 8.9\% higher. It is noteworthy that the \emph{image-only AP$_{25}$ is even higher than that achieved by CoDA with point clouds}. OV-Uni3DETR obtains 8.10\% AP$_{novel}$ and 10.23\% AP$_{all}$ using RGB images only, which surpasses CoDA by 1.56\% and 1.19\% separately using point clouds. This strongly demonstrates that our \textbf{multi-modal architecture and associated switch-modality training effectively prevent the collapse of single-modal information, thus achieving modality unifying}. Meanwhile, the integrated utilization of different modalities is further boosted. For multi-modal inference, OV-Uni3DETR achieves 15.21\% AP$_{novel}$, 8.67\% higher than previous methods. The superiority of OV-Uni3DETR can thus be further demonstrated.
%This strongly demonstrates the effectiveness of our method, since it achieves higher AP$_{25}$ using data that contains less spatial information. 

%It can also be observed that if novel class names are not available in advance during training, the 3D detection AP is even higher. This indicates that our method does not require prior information about the novel class names, which enhances its flexibility for practical applications.

Additionally, we train and evaluate OV-Uni3DETR in the same setting as OV-3DET \cite{lu2023open} on the ScanNet dataset, where 20 categories are evaluated. All categories are not annotated with 3D bounding boxes (\emph{i.e.}, there are no seen classes and all classes are novel ones). For a fair comparison, we train and evaluate our model with point clouds only, and without RGB images participating in. The AP$_{25}$ metric is listed in Tab. \ref{tab:cmpwith3ddet}. We obtain the 25.33\% AP$_{25}$, which surpasses OV-3DET by 7.31\% and CoDA by 6.01\%. This further demonstrates the effectiveness of our OV-Uni3DETR and its ability in different settings, even if no seen classes are available. For 16 classes out of the total 20 ones, we achieve the best 3D detection performance among the three methods. This validates the superiority of our method over different categories.

\begin{table}[t]
\centering
\setlength{\abovecaptionskip}{0pt}
\setlength{\belowcaptionskip}{0pt}
\caption{\textbf{The performance of OV-Uni3DETR on the KITTI and nuScenes dataset for open-vocabulary 3D object detection.} For KITTI, the car and cyclist classes are seen during training while the pedestrian class is novel. We report AP$_{25}$ with 11 recall positions on the moderate difficulty. For nuScenes, ``Car, trailer, construction vehicle, motorcycle, bicycle'' are seen and the rest five are unseen ones. }
\resizebox{\columnwidth}{!}{
\begin{tabular}{c|c|ccc|ccc|ccc}
\Xhline{1.1pt} 
\multirow{2}*{Method}& \multirow{2}*{Inputs}& \multicolumn{3}{c|}{KITTI} & \multicolumn{6}{c}{nuScenes} \\
% \cline{3-11}
 &  &AP$_{Ped.}$ &  AP$_{Car}$ & AP$_{Cyc.}$ & AP$_{novel}$ & AP$_{base}$ &  \multicolumn{1}{c@{}}{AP$_{all}$\ } & NDS$_{novel}$ & NDS$_{base}$ & NDS$_{all}$ \\ 
\hline
 Det-PointCLIP \cite{zhang2022pointclip} & P & 0.32 & 3.67 & 1.32 & 0.59 & 2.13 & 1.36 & 1.92 & 5.86 &  3.89\\
 Det-PointCLIPv2 \cite{zhu2023pointclip} & P & 0.32 & 3.58 & 1.22 & 0.61 & 2.05 & 1.33 & 1.97 & 5.74 &  3.86\\
 % Det-CLIP$^2$ \cite{zeng2023clip2} & P & \\
 3D-CLIP \cite{radford2021learning} & P+I & 1.28 & 42.28 & 21.99  & 2.74 & 12.60 & 7.67 & 8.98 & 23.81 & 16.39\\
\hline
\multirow[c]{3}{*}{OV-Uni3DETR (ours)} & P  &  19.57  & 92.44 &  56.67 & 15.48 & 61.28 & 38.39 & 15.61 & 44.71 &  30.16 \\
 & I & 9.98 & 75.14 & 18.44 & 12.54 & 55.30 & 33.93 &  14.67 & 39.43 &  27.05\\
 & P+I & \textbf{23.04} & \textbf{92.55} & \textbf{58.21} &  \textbf{18.96} & \textbf{63.34} & \textbf{41.15} & \textbf{17.05} & \textbf{46.69} &  \textbf{31.87}\\
\Xhline{1.1pt} 
\end{tabular}
}
\label{tab:outdooropen}
\end{table}

\paragraph{Outdoor 3D open-vocabulary detection.} We then evaluate on the outdoor KITTI dataset. For the simplicity of training, we do not utilize the ground-truth sampling augmentation \cite{yan2018second} here. We report the AP$_{25}$ metric with 11 recall positions on the moderate difficulty objects from the validation set, and list the results in Tab. \ref{tab:outdooropen}. To the best of our knowledge, we are the first to conduct open-vocabulary experiments on the outdoor 3D detection datasets. Outdoor point clouds are usually collected by the LiDAR sensor, where background points dominate the scene, and foreground objects are small and sparse, with significantly less points. The gap between outdoor LiDAR points and 2D images becomes larger, making detecting novel classes quite challenging in the outdoor scenes. Therefore, directly implementing CLIP-based methods to outdoor point clouds results in the limited performance. In this situation, OV-Uni3DETR still achieves 19.57\% AP$_{25}$ of the pedestrian class (the novel class) for 3D box prediction, which surpasses 3D-CLIP by 18.29\%. For multi-modal results, AP$_{novel}$ is further improved to 23.04\%. This validates the ability of our OV-Uni3DETR to detect in both indoor and outdoor scenes, thus achieving scene unifying.

For nuScenes, we report mAP and nuScenes detection score (NDS), with the match thresholds of 15 meters in Tab. \ref{tab:outdooropen}. The CBGS \cite{zhu2019class} strategy is adopted for training. For such a setting, where more categories appear in the outdoor scenes, OV-Uni3DETR achieves the performance that is equally good - 18.96\% AP$_{novel}$ and 63.34\% AP$_{base}$. It achieves the 17.05\% NDS$_{novel}$, which demonstrates that the detector can also predict object attributes like velocity well for novel classes, which is critical for the model applications in outdoor scenes.

\begin{figure}[t]
\centering
\includegraphics[width=\columnwidth]{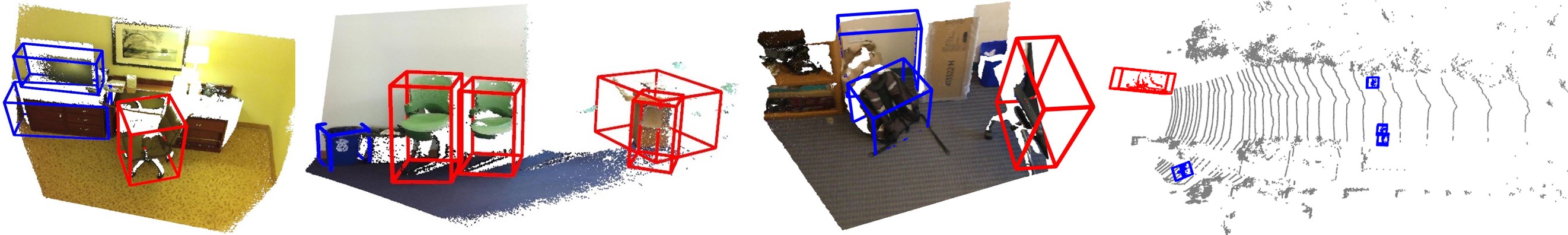}
\caption{\textbf{Visualization of OV-Uni3DETR for open-vocabulary 3D detection}  on the SUN RGB-D (the first and the second), ScanNet (the third) and KITTI (the fourth) dataset. The red boxes are {\color{red} base} classes and blue boxes are {\color{blue} novel} classes.}
\label{fig:visual}
\end{figure}

\paragraph{Visualization.} We provide visualized results in Fig. \ref{fig:visual}, where novel class objects are in blue boxes. As can be seen, OV-Uni3DETR well recognizes and localizes novel classes in indoor and outdoor scenes. This further validates its ability.

\begin{table}[t]
\centering
% {\fontsize{6.5pt}{7.8pt}\selectfont
\begin{minipage}[c]{0.43\columnwidth}
\centering
\setlength{\abovecaptionskip}{-2pt}
\setlength{\belowcaptionskip}{2pt}
\caption{\textbf{The 2D detection performance of OV-Uni3DETR on SUN RGB-D} compared to 2D detectors with image-text pre-training. }
    \resizebox{0.99\columnwidth}{!}{
    \begin{tabular}{c|c|ccc}
    \Xhline{1.1pt} 
    Method & Inputs & AP$_{novel}^{2D}$ & AP$_{base}^{2D}$ & AP$_{all}^{2D}$ \\ 
    \hline
    CLIP \cite{jia2021scaling} & I & 0.52 & 11.78 & 2.97\\
    RegionCLIP \cite{zhong2022regionclip} & I & 2.24 & 18.35 & 5.74\\
    \hline
    \multirow[c]{3}{*}{\shortstack{OV-Uni3DETR \\ (ours)}} & P & 0.40 & 3.05 & 0.98\\
     & I &  5.21 & 19.25 & 8.26\\
     & P+I &  \textbf{6.52} & \textbf{19.68} & \textbf{9.38}\\
    \Xhline{1.1pt} 
    \end{tabular}
    }
    \label{tab:sunrgbdopen2d}
\end{minipage}
\hfill
\begin{minipage}[c]{0.54\columnwidth}
\setlength{\abovecaptionskip}{0pt}
\setlength{\belowcaptionskip}{0pt}
\centering
\caption{\textbf{The performance of OV-Uni3DETR on SUN RGB-D for closed-vocabulary monocular 3D detection.} The AP$^{3D}$ metric is adopted from OMNI3D \cite{brazil2023omni3d}. }
    \resizebox{\columnwidth}{!}{
    \begin{tabular}{c|c|c|ccc}
    \Xhline{1.1pt} 
    Method & Backbone & Trained on & AP$_{25}$ & AP$_{50}$ & AP$^{3D}$ \\ 
    \hline
     ImVoxelNet \cite{rukhovich2022imvoxelnet}&\multirow[c]{4}{*}{ResNet34} & SUN RGB-D & 34.1 & 12.8 & 30.6\\
     Cube RCNN \cite{brazil2023omni3d} & & SUN RGB-D & - & - & 34.7\\
     Cube RCNN & & OMNI3D$_{\rm IN}$ & - & -  & 35.4\\
     OV-Uni3DETR (ours) & & SUN RGB-D & \textbf{41.6} & \textbf{15.4} &  \textbf{37.7 }\\
    \hline
    ImVoxelNet & \multirow[c]{2}{*}{ResNet50} & SUN RGB-D & 40.9 &  13.5 & 36.3\\
     OV-Uni3DETR (ours) & & SUN RGB-D & \textbf{44.6} & \textbf{16.1}  & \textbf{39.7}\\
    \Xhline{1.1pt} 
    \end{tabular}
    }
    \label{tab:sunrgbdclosed}
\end{minipage}
\end{table}

\paragraph{2D open-vocabulary detection.} Besides the 3D boxes, OV-Uni3DETR predicts the 4-dim 2D boxes. Here we compare its 2D performance with the  Faster RCNN \cite{ren2015faster} detector trained on the same SUN RGB-D dataset. To recognize novel classes, we initialize Faster RCNN with CLIP \cite{jia2021scaling} and RegionCLIP \cite{zhong2022regionclip} parameters for comparison. With only RGB images as input, we achieve the 5.21\% AP$_{novel}$ for the 2D boxes, which is 2.97\% higher than RegionCLIP. Although point cloud data struggle to produce detection results that are comparable to that from RGB only, partially because of the lack of 2D information, they can contribute to further improvement for multi-modal inference. We achieve the 6.52\% AP$_{novel}$ for the multi-modal performance, 4.28\% higher than RegionCLIP. This demonstrates that OV-Uni3DETR can also predict 2D boxes well.

\subsection{Closed-Vocabulary 3D Object Detection}

In this subsection, we evaluate in the traditional closed-vocabulary 3D detection setting. As the point cloud branch is adopted from Uni3DETR, here we evaluate with only RGB images as input, and compare with existing RGB-only methods.

\paragraph{Indoor 3D closed-vocabulary detection.} We first evaluate on SUN RGB-D for the 10 category setting from VoteNet \cite{qi2019deep}, and list 3D AP in Tab. \ref{tab:sunrgbdclosed}. With the ResNet34 backbone, OV-Uni3DETR obtains 37.7\% AP$^{3D}$, which surpasses ImVoxelNet by 7.1\% and Cube RCNN by 3\%. Besides, it is 2.3\% higher than Cube RCNN trained on OMNI3D$_{\rm IN}$, with significantly more RGB images participating in training. With the ResNet50 backbone, the AP$^{3D}$ is further improved to 39.7\%, 3.4\% higher than ImVoxelNet. This illustrates that OV-Uni3DETR can also achieve an excellent performance for the close-vocabulary setting.

\begin{table}[t]
\centering
\begin{minipage}[c]{0.52\columnwidth}
\setlength{\abovecaptionskip}{0pt}
\setlength{\belowcaptionskip}{0pt}
\centering
\caption{\textbf{The performance of OV-Uni3DETR on KITTI test for closed-vocabulary monocular 3D object detection.} *: AP$^{3D}$ on the moderate car is the most important metric. $^\S$: the method uses more images (OMNI3D) for training.}
\resizebox{\columnwidth}{!}{
\begin{tabular}{c|ccc|ccc}
\Xhline{1.1pt} 
\multirow{2}*{Method} & \multicolumn{3}{c|}{AP$^{3D}$} & \multicolumn{3}{c}{AP$^{BEV}$} \\
% \cline{2-7}
  & Easy & Mod.* & Hard & Easy & Mod. & Hard \\ 
\hline
 SMOKE \cite{liu2020smoke} & 14.03 & 9.76 & 7.84 & 20.83 & 14.49 & 12.75\\
 PGD \cite{wang2022probabilistic} & 19.05 & 11.76 & 9.39 & 26.89 & 16.51 & 13.49\\
 MonoRCNN \cite{shi2021geometry} & 18.36 & 12.65 & 10.03 & 25.48 & 18.11 & 14.10\\
 MonoFlex \cite{zhang2021objects} & 19.94 & 13.89 & 12.07 & 28.23 & 19.75 & 16.89 \\
 GUPNet \cite{lu2021geometry} & 20.11 & 14.20 & 11.77 & - & - & -\\
 Cube RCNN $^\S$ \cite{brazil2023omni3d} & 23.59 & 15.01 & 12.56 & 31.70 & 21.20 & 18.43 \\
 OV-Uni3DETR (ours) & 20.81 & \textbf{15.95} & \textbf{13.95} & 29.80 & \textbf{22.74} & \textbf{20.20}\\
\Xhline{1.1pt} 
\end{tabular}
}
\label{tab:kitticlosed}
\end{minipage}
\hfill
\begin{minipage}[c]{0.45\columnwidth}
\centering
\setlength{\abovecaptionskip}{0pt}
\setlength{\belowcaptionskip}{2pt}
\caption{\textbf{Effect of cycle-modality propagation on the SUN RGB-D dataset.} 2D$\rightarrow$3D and 3D$\rightarrow$2D knowledge propagation are studied when RGB images exist. }
    \resizebox{\columnwidth}{!}{
    \begin{tabular}{c|cc|ccc}
    \Xhline{1.1pt} 
    Inputs & 2D$\rightarrow$3D & 3D$\rightarrow$2D & AP$_{novel}$ & AP$_{base}$ & AP$_{all}$ \\ 
    \hline
    \multirow[c]{4}{*}{I} &  &  & N/A & 28.12 & N/A\\
      & \ding{51}  &  & 3.86 & 28.17 & 9.14\\
      &  & \ding{51} &  2.02 & 29.24 & 7.94\\
      & \cellcolor{mycolor}{\ding{51}} & \cellcolor{mycolor}{\ding{51}}  & \cellcolor{mycolor}{\textbf{5.41}} & \cellcolor{mycolor}{\textbf{29.51}} & \cellcolor{mycolor}{\textbf{10.65}}\\
    \hline
    \multirow[c]{4}{*}{P+I} &  &  &  N/A & 48.45 & N/A\\
      & \ding{51} &  & 11.21 & 48.42 & 19.30\\
      &  & \ding{51} & 3.72 & 49.23 & 13.61\\
      & \cellcolor{mycolor}{\ding{51}} & \cellcolor{mycolor}{\ding{51}} & \cellcolor{mycolor}{\textbf{12.96}} & \cellcolor{mycolor}{\textbf{49.25}} & \cellcolor{mycolor}{\textbf{20.85}}\\
    \Xhline{1.1pt} 
    \end{tabular}
    }
    \label{tab:2dimage}
\end{minipage}
\end{table}

\paragraph{Outdoor 3D closed-vocabulary detection.} We then evaluate on KITTI and compare it with previous monocular 3D detection methods on the test set. We list the AP$_{70}$ metric with 40 recall positions in Tab. \ref{tab:kitticlosed}. We obtain the 15.95\% AP for the moderate difficulty, which surpasses MonoFlex by 2.06\% and GUPNet by 1.75\%. Compared with Cube RCNN, which is trained on the OMNI3D benchmark with more images, the 3D detection AP is 0.94\% higher. The ability in the outdoor closed-vocabulary setting can thus be validated.

\subsection{Ablation Study}

Finally, we conduct ablation studies in this subsection. We mainly analyze the effect of the cycle-modality propagation and multi-modal learning.

\paragraph{Cycle-modality propagation.} Tab. \ref{tab:2dimage} analyzes the effect of the cycle-modality propagation in open-vocabulary learning. We conduct such a study when RGB images exist, since the 3D$\rightarrow$2D propagation requires images as input. Without the cycle-modality propagation, the detector does not have the open-vocabulary ability, thus cannot detect novel class objects. The \emph{2D$\rightarrow$3D propagation helps leverage semantic knowledge in the 2D domain, thus helps expand the 3D vocabulary size}. The detector can thus perform open-vocabulary detection. The 3D$\rightarrow$2D propagation further helps improve the performance. When only images are available, AP$_{novel}$ is enhanced by 1.55\%, and AP$_{all}$ is increased by 1.51\%. Meanwhile, the multi-modal AP$_{novel}$ is improved by 1.75\%. The \emph{3D$\rightarrow$2D propagation introduces geometric knowledge into 2D detection images, where more annotated categories and diversified contained scenes boost open-vocabulary learning}. Through our proposed cycle-modality propagation, the 2D semantic and 3D geometric knowledge can be leveraged, which contributes to the comprehensive utilization of knowledge from different modalities. Therefore, OV-Uni3DETR can perform open-vocabulary 3D detection with modality unifying.

\begin{table}[t]
\centering 
\setlength{\abovecaptionskip}{0pt}
\setlength{\belowcaptionskip}{0pt}
\caption{\textbf{Effect of the details in multi-modal learning on SUN RGB-D.} ``Switched'', ``3Dconv'', and ``dualb'' are short for the switched-modality training strategy, 3D convolution with BN for feature regularization, and the dual-branch structure. These designs reduce the strong dependency on the point cloud modality. The image-only AP can thus be on par with the single-modality baseline. }
    \resizebox{0.95\columnwidth}{!}{
    \begin{tabular}{c|ccc|ccc|ccc|ccc}
    \Xhline{1.1pt} 
      & \multirow[c]{2}{*}{switched} & \multirow[c]{2}{*}{\makecell{3Dconv}} & \multirow[c]{2}{*}{dualb} & \multicolumn{3}{c|}{AP$_{novel}$} & \multicolumn{3}{c|}{AP$_{base}$} & \multicolumn{3}{c}{AP$_{all}$}\\
      % \cline{5-13}
      &  &  &  & P & \cellcolor{mycolor}{I} & P+I &  P & \cellcolor{mycolor}{I} & P+I &  P & \cellcolor{mycolor}{I} & P+I  \\ 
    \hline
    single-modality &  &  &  & 9.64 & \cellcolor{mycolor}{5.35} & - & 47.83 & \cellcolor{mycolor}{30.15} & -  & 17.94 & \cellcolor{mycolor}{10.74} & -\\
    \hline 
    \multirow[c]{4}{*}{multi-modality} &  & \ding{51} & \ding{51} & 6.36 & \cellcolor{mycolor}{1.12} & 12.04  & 43.27 & \cellcolor{mycolor}{10.74} & 47.89  & 14.38 & \cellcolor{mycolor}{3.21} & 19.81\\
    & \ding{51} &  & \ding{51} & 9.57 & \cellcolor{mycolor}{3.37} & 10.92 & 47.86 & \cellcolor{mycolor}{25.82} & 49.01 & 15.54 & \cellcolor{mycolor}{8.25} & 19.20\\
    & \ding{51} & \ding{51} & &  6.34 & \cellcolor{mycolor}{4.41} & 9.65 & 41.30 & \cellcolor{mycolor}{26.73} & 42.84 & 13.94 & \cellcolor{mycolor}{9.26} & 16.86\\
    & \ding{51} & \ding{51} & \ding{51} &  \textbf{9.66} & \cellcolor{mycolor}{\textbf{5.41}} & \textbf{12.96} & \textbf{48.29} & \cellcolor{mycolor}{\textbf{29.51}} & \textbf{49.25} & \textbf{18.06} & \cellcolor{mycolor}{\textbf{10.65}} & \textbf{20.85}\\
    \Xhline{1.1pt} 
    \end{tabular}
    }
    \label{tab:structureablation}
\end{table}

\paragraph{Multi-modal learning.} Tab. \ref{tab:structureablation} analyzes the effect of our designed multi-modal learning manner. When either point clouds or RGB images are used for inference, the AP$_{25}$ metrics are basically equal to or surpass those from single-modality training. Without switched-modality training, the detection performance deteriorates severely for single-modality inference. Especially when only images are available, switched-modality training contributes to the AP$_{novel}$ improvement from 1.12\% to 5.41\%. By random switching, the model can accept image-only features during training, thus preventing image features from being ignored and reducing the strong dependency on point clouds. Equally, 3D convolution with BN helps improve the image-only AP$_{novel}$ from 3.37\% to 5.41\%. The reason is that 3D convolution with batch normalization regularizes different modality features, reducing the modality gap between the point cloud features and the image features. This prevents feature discrepancy and the suppression of image features compared to more information-abundant point cloud features. Furthermore, the dual-branch structure helps distinguish information flows of different modalities, thus avoiding the effects of generated noisy 3D boxes. It thus helps improve the performance with any modality inputs: point cloud only AP$_{novel}$ from 6.34\% to 9.66\%, and image-only AP$_{novel}$ from 4.41\% to 5.41\%. The capacity of our multi-modal learning for modality unifying can be demonstrated.

\section{Conclusion}

In this paper, we mainly propose OV-Uni3DETR, a unified multi-modal open-vocabulary 3D detector. With the help of multi-modal learning and the cycle-modality knowledge propagation, our OV-Uni3DETR recognizes and localizes novel classes well, achieving modality unifying and scene unifying. Experiments demonstrate its strong ability in both open-vocabulary and close-vocabulary settings, both indoor and outdoor scenes, and with any modality data inputs. Addressing unified open-vocabulary 3D detection in the multi-modal context, we believe our research will stimulate following research along the promising but challenging universal 3D computer vision direction.

\section*{Acknowledgements}

This work is supported by the state key development program in 14th Five-Year under Grant No. 2021YFF0602102-2, the National Natural Science Foundation of China No. 62201484, HKU Startup Fund, and HKU Seed Fund for Basic Research. We also thank the research fund under Grant No. 20222910018 from the Huachengzhiyun Soft Co., Ltd.

\appendix

% \section{How does OV-Uni3DETR distinguish?}
\section{Overview of 3D Object Detection Research State}

\vspace{-15pt}
\begin{table}[h]
\centering
\setlength{\abovecaptionskip}{0pt}
\setlength{\belowcaptionskip}{0pt}
\caption{\textbf{The overview of existing 3D object detectors on their capability of scene unifying, modality unifying and category unifying.} Scene unifying mainly includes detecting in indoor and outdoor scenes. For modality unifying, the main aspects include utilizing both point clouds and images for training ("P-train", "I-train"), and being modality-switchable during inference ("switch"). For category unifying, the main process is about open-vocabulary learning ("open").}
\resizebox{0.99\columnwidth}{!}{
\begin{tabular}{ccc|cc|ccc|cc}
\Xhline{1.1pt} 
\multicolumn{3}{c|}{\multirow{2}*{Methods}} & \multicolumn{2}{c|}{Scene} & \multicolumn{3}{c|}{Modality} & \multicolumn{2}{c}{Category} \\
& &  & indoor & outdoor & P-train & I-train & switch & closed & open \\
% \cline{3-8}
\hline
\multirow{4}*{\shortstack{indoor}} & VoteNet \cite{qi2019deep} & ICCV'19 & \ding{51} & \ding{55} & \ding{51} & \ding{55} & \ding{55} & \ding{51} & \ding{55} \\
 & 3DETR \cite{misra2021end} & ICCV'21 & \ding{51} & \ding{55} & \ding{51} & \ding{55} & \ding{55} & \ding{51} & \ding{55} \\
 & FCAF3D \cite{rukhovich2022fcaf3d} & ECCV'22 & \ding{51} & \ding{55} & \ding{51} & \ding{55} & \ding{55} & \ding{51} & \ding{55} \\
 & NeRF-Det \cite{xu2023nerf} & ICCV'23 & \ding{51} & \ding{55} & \ding{55} & \ding{51} & \ding{55} & \ding{51} & \ding{55} \\
\hline 
\multirow{5}*{\shortstack{outdoor}} & PointPillars \cite{lang2019pointpillars}  & CVPR'19 & \ding{55} & \ding{51} & \ding{51} & \ding{55} & \ding{55} & \ding{51} & \ding{55} \\
 & CenterPoint \cite{yin2021center} & CVPR'21 & \ding{55} & \ding{51} & \ding{51} & \ding{55} & \ding{55} & \ding{51} & \ding{55} \\
 & MonoFlex \cite{zhang2021objects} & CVPR'21 & \ding{55} & \ding{51} & \ding{55} & \ding{51} & \ding{55} & \ding{51} & \ding{55} \\
 & BEVFormer \cite{li2022bevformer} & ECCV'22 & \ding{55} & \ding{51} & \ding{55} & \ding{51} & \ding{55} & \ding{51} & \ding{55} \\
 & VoxelNeXt \cite{chen2023voxelnext} & CVPR'23 & \ding{55} & \ding{51} & \ding{51} & \ding{55} & \ding{55} & \ding{51} & \ding{55} \\
\hline
\multirow{5}*{\shortstack{multi-modal}} &  MVXNet \cite{sindagi2019mvx} & ICRA'19 & \ding{55} & \ding{51} & \ding{51} & \ding{51} & \ding{55} & \ding{51} & \ding{55} \\
 & ImVoteNet \cite{qi2020imvotenet} & CVPR'20 & \ding{51} & \ding{55} & \ding{51} & \ding{51} & \ding{55} & \ding{51} & \ding{55} \\
 & TokenFusion \cite{wang2022multimodal} & CVPR'22 & \ding{51} & \ding{55} & \ding{51} & \ding{51} & \ding{55} & \ding{51} & \ding{55} \\
 & UVTR \cite{li2022unifying} & NeurIPS'22 & \ding{55} & \ding{51} & \ding{51} & \ding{51} & \ding{55} & \ding{51} & \ding{55} \\
 & VirConvNet \cite{wu2023virtual} & CVPR'23 & \ding{55} & \ding{51} & \ding{51} & \ding{51} & \ding{55} & \ding{51} & \ding{55} \\
\hline
\multirow{4}*{\shortstack{scene-unified}} & EPNet \cite{huang2020epnet} & ECCV'20 & \ding{51} & \ding{51} & \ding{51} & \ding{51} & \ding{55} & \ding{51} & \ding{55} \\
 & ImVoxelNet \cite{rukhovich2022imvoxelnet}  & WACV'22 & \ding{51} & \ding{51} & \ding{55} & \ding{51} & \ding{55} & \ding{51} & \ding{55} \\
 & Cude RCNN \cite{brazil2023omni3d}  & CVPR'23 & \ding{51} & \ding{51} & \ding{55} & \ding{51} & \ding{55} & \ding{51} & \ding{55} \\
 & Uni3DETR \cite{wang2023uni3detr} & NeurIPS'23 & \ding{51} & \ding{51} & \ding{51} & \ding{55} & \ding{55} & \ding{51} & \ding{55} \\
\hline
\shortstack{modality-unified} & MetaBEV \cite{Ge_2023_ICCV} & ICCV'23  & \ding{55} & \ding{51} & \ding{51} & \ding{51} & \ding{51} & \ding{51} & \ding{55} \\
\hline
\multirow{2}*{\shortstack{open-vocabulary} } & OV-3DET \cite{lu2023open} & CVPR'23 & \ding{51} & \ding{55} & \ding{51} & \ding{55} & \ding{55} & \ding{51} & \ding{51}\\
 & CoDA \cite{cao2023coda} & NeurIPS'23 & \ding{51} & \ding{55} & \ding{51} & \ding{55} & \ding{55} & \ding{51} & \ding{51}\\
\Xhline{0.8pt} 
\multicolumn{3}{c|}{\cellcolor{mycolor}{ {\color{red} $\star$} OV-Uni3DETR (ours) }}  & \cellcolor{mycolor}{\ding{51}} & \cellcolor{mycolor}{\ding{51}} & \cellcolor{mycolor}{\ding{51}} & \cellcolor{mycolor}{\ding{51}} & \cellcolor{mycolor}{\ding{51}} & \cellcolor{mycolor}{\ding{51}} & \cellcolor{mycolor}{\ding{51}}\\
\Xhline{1.1pt} 
\end{tabular}
}
\label{tab:overview}
\end{table}
\vspace{-5pt}

We list the overview of existing 3D object detectors about their unifying capability in Tab. \ref{tab:overview}. Related research on 3D universal detection still remains lagging behind that of 2D. A universal 3D detector, we believe, should at least have the ability to utilize any modality data to detect any category in any scene. As such, it should achieve unifying across three levels: scene, modality, and category. Early research primarily focuses on the specific domains. Some recent works achieve unifying on one aspect, but fail to address all three ones. In comparison, OV-Uni3DETR performs open-vocabulary 3D detection with modality unifying and scene unifying, and well achieves the targets of unifying along the modality, scene, category levels. Therefore, it greatly advances existing research towards the goal of universal 3D object detection. We believe OV-Uni3DETR can become a significant step towards the future of 3D foundation models.

\section{More Implemental Details}

\paragraph{SUN RGB-D}  \cite{song2015sun}.  For training OV-Uni3DETR on the SUN RGB-D dataset, we adopt the initial learning rate of 2e-5 and the batch size of 32 for 90 epochs, and the learning rate is decayed by 10x on the 70th and 80th epoch. For point clouds, we filter the input point clouds in the range [-3.2m, 3.2m] for the $x$ axis, [-0.2m, 6.2m] for the $y$ axis and [-2m, 0.56m] for the $z$ axis. We randomly flip the point clouds along the $x$ axis and randomly sample 20,000 points for data augmentation. Common global translation, rotation and scaling strategies are also adopted here. For RGB images, we only adopt pixel-level data augmentation strategies, including brightness, contrast, saturation, and channel swapping. No spatial-level augmentations are utilized.

\paragraph{ScanNet}  \cite{dai2017scannet}. For the ScanNet dataset, we adopt the totally same setting as \cite{cao2023coda} for training and inference, where point clouds are generated from the single-view depth images. We first normalize the point clouds to the center of the view, then adopt the range of [-5.12m, 5.12m] for the $x$ and $y$ axis and [-1.28m, 1.28m] for the $z$ axis. We train OV-Uni3DETR with the initial learning rate of 1e-5 and the batch size of 24 for 40 epochs, and the learning rate is decayed by 10x on the 32nd and 38th epoch. As we can see, the training epochs we require is significantly less than the previous method CoDA \cite{cao2023coda}. 

Besides the experimental setting in our main paper, we also utilize a multi-view setting of the ScanNet dataset in this supplementary material. Specifically, each scene consists of plenty of multi-view images, and the point clouds are reconstructed from the multiple multi-view images for a panoramic, full-angle large scene. We also choose the top 10 classes for base classes, and the other 50 ones for novel classes. For point clouds, we adopt the range of [-6.4m, 6.4m] for the $x$ and $y$ axis and [-0.1m, 2.46m] for the $z$ axis after global alignment. The input point clouds are randomly flipped along both the $x$ and $y$ axis. For RGB images, we randomly select 8 multi-view images for training. For inference, 24 input views are randomly selected for the multi-modal inference and 40 input views are utilized for RGB-only inference. Other hyper-parameters and operations are the same as the SUN RGB-D dataset.

\paragraph{KITTI} \cite{geiger2013vision}. For the KITTI dataset, we do not use ground-truth sampling and the object-level noise strategy augmentations during training. Instead, we randomly flip the point clouds along the $x$ axis and randomly sample 18,000 points for data augmentation. For RGB images, besides the pixel-level data augmentation strategies, we also randomly flip the input images along the $x$ axis.

\paragraph{nuScenes} \cite{caesar2020nuscenes}. Compared to KITTI, the nuScenes dataset covers a larger range, with 360 degrees around the LiDAR instead of only the front view. Each scene consists of 6 multi-view images. We train OV-Uni3DETR for 20 epochs with the cyclic schedule. The CBGS \cite{zhu2019class} strategy is adopted for training point clouds. %Other augmentation strategies are basically the same as that for KITTI.

\section{More Method Details}

\vspace{-10pt}

\begin{algorithm}%[t]
\caption{The overall procedure for training OV-Uni3DETR.}
\begin{algorithmic}
\State \hspace*{-0.18in} \textbf{Input}:
\State 1. 3D data ($X_P$, $X_I$): point clouds and corresponding 3D detection images, inherently \hspace*{0.1in} associated with camera parameters $K$, $R_t$ and 3D annotations $\{c_i, bb_i^{3D}, bb_i^{2D}\}$.
\State 2. 2D data $X_I^{2D}$: 2D detection images inherently associated with 2D labels $\{c_i, bb_i^{2D}\}$, \hspace*{0.1in} pre-trained 2D detector $\Phi^{2D}$.
% \State 3. Pre-trained 2D open-vocabulary detector $\Phi^{2D}$.
\State \hspace*{-0.18in} \textbf{Training}:
\State \textbf{2D $\rightarrow$ 3D knowledge propagation}:
\State \hspace*{0.1in} 1. Apply $\Phi^{2D}$ on $X_I$: $\{\hat{c}_i, \hat{bb}_i^{2D}\} = \Phi^{2D}(X_I)$.
\State \hspace*{0.1in} 2. Project 2D boxes to 3D for 3D boxes: $\hat{bb}_i^{3D} = \hat{bb}_i^{2D}\circ(KR_t)^{-1}$
\State \textbf{3D $\rightarrow$ 2D knowledge propagation}:
\State \hspace*{0.1in} 1. Pre-train a class-agnostic 3D detector $\Phi^{3D}_{CA}$ with the module for camera extrinsic \hspace*{0.23in} parameter prediction using 3D detection images $X_I$, $R_t$, $\{bb_i^{3D}\}$.
\State \hspace*{0.1in} 2. Apply $\Phi^{3D}_{CA}$ on $X_I^{2D}$: $\hat{R_t}, \{\hat{bb}_i^{2D}, \hat{bb}_i^{3D}\} = \Phi^{3D}_{CA}(X_I^{2D})$.
\State \hspace*{0.1in} 3. Set camera intrinsic parameters $K$ according to image sizes for $X_I^{2D}$.
\State \hspace*{0.1in} 4. Obtain class-specific 3D boxes $\{\hat{c}_i, {bb}_i^{2D}, \hat{bb}_i^{3D}\}$ for $X_I^{2D}$ by Hungarian matching.
% \For{$i=1$; $i<=N$; $i++$}
% \State Randomly selecting $(X_P, X_I)$ or $X_I^{2D}$ as the training sample.
\State \textbf{Network forwarding}:
\State \hspace*{0.1in} 1. Extract point cloud features $F_P$, image features $F_I$, and project image features  \hspace*{0.25in} to the voxel space through $KR_t$ for $F_I'$.
\State \hspace*{0.1in} 2. Obtain multi-modal features: $F_M = {\rm BN}({\rm conv}(F_P)) + {\rm BN}({\rm conv}(F_I')$).
\State \hspace*{0.1in} 3. Randomly select $F_P$, $F_I'$, $F_M$ for category, 3D box and 2D box prediction.
\State \hspace*{0.1in} 4. Supervise with $\{c_i, bb_i^{3D}, bb_i^{2D}\} + \{\hat{c}_i, \hat{bb}_i^{3D}, \hat{bb}_i^{2D}\}$ for training.
% \EndFor
% \State \hspace*{-0.18in} \textbf{Testing}:
% \State Ensemble testing results from all models to generate ultimate results.
\end{algorithmic}
\label{alg:process}
\end{algorithm}
\vspace{-5pt}

We summarize our method in Algorithm \ref{alg:process}. Specifically, OV-Uni3DETR utilizes various sources of data, including point clouds, 3D detection images and 2D detection images for training. 3D detection images align with point clouds and are 3D box annotated, while 2D detection images are not aligned and only 2D box annotated. 

For training OV-Uni3DETR, we introduce the concept of cycle-modality propagation. Specifically, for 2D $\rightarrow$ 3D knowledge propagation, we leverage a pre-trained 2D detector on 3D detection images, and project the generated 2D boxes to the 3D space to propagate semantic knowledge into the 3D domain. For 3D $\rightarrow$ 2D knowledge propagation, we first pre-train a class-agnostic 3D detector with a camera extrinsic parameter prediction branch, and apply it on 2D detection images to propagate geometric knowledge into the 2D domain.

With knowledge propagation, the input data are forwarded into the network for training, and the generated boxes from cycle-modality propagation are integrated into the ground-truth to supervise the training. The switch-modality training strategy is utilized for modality unifying.

\section{More Quantitative Results}

\subsection{ScanNet Multi-View Setting}

\begin{table}[t]
\centering
\begin{minipage}[c]{0.49\columnwidth}
\setlength{\abovecaptionskip}{0pt}
\setlength{\belowcaptionskip}{0pt}
\caption{\textbf{The performance of OV-Uni3DETR on the ScanNet multi-view setting for open-vocabulary 3D object detection.} P denotes the point cloud inputs and I denotes image inputs.}
\resizebox{\columnwidth}{!}{
\begin{tabular}{c|c|ccc}
\Xhline{1.1pt} 
Method & Inputs & AP$_{novel}$ & AP$_{base}$ & AP$_{all}$ \\ 
\hline
 Det-PointCLIP \cite{zhang2022pointclip} & P & 0.07 & 1.05 & 0.23\\
 Det-PointCLIPv2 \cite{zhu2023pointclip} & P  & 0.06 & 1.01 & 0.22\\
 % Det-CLIP$^2$ \cite{zeng2023clip2} & P & 0.14 & 1.76 & 0.40\\
 3D-CLIP \cite{radford2021learning} & P+I & 2.52 & 11.21 & 3.97 \\
 % CoDA \cite{cao2023coda} & P & 6.54 & 21.57 & 9.04\\
\hline
\multirow[c]{3}{*}{OV-Uni3DETR (ours)} & P &  10.59  & 44.39 & 16.22\\
 & I & 7.70 & 28.39  & 11.15 \\
 & P+I & \textbf{13.72} & \textbf{48.05}  & \textbf{19.44}\\
\Xhline{1.1pt} 
\end{tabular}
}
\label{tab:scannetopen}
\end{minipage}
\hfill
\begin{minipage}[c]{0.49\columnwidth}
\centering
\setlength{\abovecaptionskip}{0pt}
\setlength{\belowcaptionskip}{1pt}
\caption{\textbf{The performance of OV-Uni3DETR on the ScanNet dataset for closed-vocabulary 3D object detection.} Only multi-view RGB images are used for training and inference.}
    \resizebox{\columnwidth}{!}{
    \begin{tabular}{c|c|c|c}
    \Xhline{1.1pt} 
    Method & \# Training & \# Inference & AP$_{25}$ \\ 
    \hline
     \multirow[c]{2}{*}{ImVoxelNet \cite{rukhovich2022imvoxelnet}} & \multirow[c]{2}{*}{20 views} & 20 views &  44.1 \\
      &  & 50 views & 48.1\\
    \hline 
    \multirow[c]{4}{*}{NeRF-Det \cite{xu2023nerf}} & \multirow[c]{2}{*}{20 views} & 20 views &  44.9\\
      &  & 50 views & 48.5\\
      \cline{2-4}
      & \multirow[c]{2}{*}{50 views} & 50 views & 51.3\\
       &  & 100 views &  52.3\\
      \hline
    ImGeoNet \cite{tu2023imgeonet} & 50 views & 50 views & 54.8\\
      \hline 
    \multirow[c]{2}{*}{ OV-Uni3DETR (ours)} & \multirow[c]{2}{*}{20 views} & 20 views & \textbf{52.3} \\
      &  & 50 views & \textbf{55.1}\\
    \Xhline{1.1pt} 
    \end{tabular}
    }
    \label{tab:scannetclosed}
\end{minipage}
\end{table}

\paragraph{3D open-vocabulary detection.} We evaluate OV-Uni3DETR on the ScanNet dataset, multi-view setting, and list the AP$_{25}$ metric in Tab. \ref{tab:scannetopen}. We randomly select 8 multi-view images from a single scene for training. For inference, 24 input views are randomly selected for the multi-modal inference and 40 input views are utilized for RGB-only inference. We observe that the superiority of our method equally becomes remarkable on the ScanNet dataset. We achieve 10.59\% AP$_{novel}$ using only point cloud data, and 13.72\% AP$_{novel}$ using multi-modal data. The listed AP here is a little lower than that in our main paper for the ScanNet single-view setting, because the ScanNet scene here is reconstructed from multi-view images, thus covers a larger range and contains more objects. The performance of our OV-Uni3DETR in such the setting further demonstrates its ability for utilizing multi-view images.

\paragraph{3D closed-vocabulary detection.} We then evaluate OV-Uni3DETR on the ScanNet dataset for the 18 category closed-vocabulary setting. With 20-view images participating in training and inference, our OV-Uni3DETR obtains the 52.3\% AP$_{25}$, which surpasses NeRF-Det by 7.4\% under the same conditions. When the images for inference increase to 50 views, our detector further achieves the 55.1\% AP$_{25}$,  6.6\% higher than NeRF-Det. The 3D detection performance is even higher than ImGeoNet, which uses more images (50-view) for training. This demonstrates that for the multi-view close-vocabulary indoor detection setting, OV-Uni3DETR can also obtain the state-of-the-art performance.

\subsection{Ablation Study on Training Data}

% % \vspace{-15pt}
% \begin{table}
% \centering 
% \setlength{\abovecaptionskip}{0pt}
% \setlength{\belowcaptionskip}{0pt}
% \caption{\textbf{Ablation study on the SUN RGB-D dataset about the 2D detection images utilized.} The format of table entries is: dataset name - training vocabulary number. Only RGB images are used for inference.}
%     \resizebox{0.5\columnwidth}{!}{
%     \begin{tabular}{c|ccc}
%     \Xhline{1.1pt} 
%     2D detection images & AP$_{novel}$ & AP$_{base}$ & AP$_{all}$ \\ 
%     \hline
%     % sunrgbd-46c & 10.07 & 48.36 & 18.39 \\
%     coco-80c & 3.29 & 32.19 & 9.57\\
%     scannet-100c & 5.34 & 33.36 & 11.43\\
%     scannet-200c & 5.65 & 28.53 & 10.62\\
%     objects365-365c & 5.41 & 29.51 & 10.65\\
%     openimages-500c & 5.79 & 32.61 & 11.62\\
%     lvis-1230c & 5.89 & 31.03 & 11.35\\
%     \Xhline{1.1pt} 
%     \end{tabular}
%     }
%     \label{tab:data}
% \end{table}

\begin{table}[t]
\centering
\begin{minipage}[c]{0.52\columnwidth}
\setlength{\abovecaptionskip}{0pt}
\setlength{\belowcaptionskip}{0pt}
\centering
\caption{\textbf{Ablation study on the SUN RGB-D dataset about the 2D detection images utilized.} The format of table entries is: dataset name - training vocabulary number. Only RGB images are used for inference.}
    \resizebox{1.0\columnwidth}{!}{
    \begin{tabular}{c|ccc}
    \Xhline{1.1pt} 
    2D detection images & AP$_{novel}$ & AP$_{base}$ & AP$_{all}$ \\ 
    \hline
    % sunrgbd-46c & 10.07 & 48.36 & 18.39 \\
    coco-80c & 3.29 & 32.19 & 9.57\\
    scannet-100c & 5.34 & 33.36 & 11.43\\
    scannet-200c & 5.65 & 28.53 & 10.62\\
    objects365-365c & 5.41 & 29.51 & 10.65\\
    openimages-500c & 5.79 & 32.61 & 11.62\\
    lvis-1230c & 5.89 & 31.03 & 11.35\\
    \Xhline{1.1pt} 
    \end{tabular}
    }
    \label{tab:data}
\end{minipage}
\hfill
\begin{minipage}[c]{0.45\columnwidth}
\centering
\setlength{\abovecaptionskip}{0pt}
\setlength{\belowcaptionskip}{0pt}
\caption{\textbf{The performance of the class-agnostic 3D detector on the SUN RGB-D dataset.} The class-agnostic 3D detector is used in 2D $\rightarrow$ 3D propagation and we compare it with the class-specific baseline.}
\resizebox{1.0\columnwidth}{!}{
\begin{tabular}{c|ccc}
\hline
 &  AR$_{novel}$ & AR$_{base}$ & AR$_{all}$ \\ 
\hline
class-specific &  35.03 & 69.08 & 43.24 \\
class-agnostic & 55.69 & 88.20 & 68.87 \\
\hline 
\end{tabular}
}
\label{tab:ca}
\end{minipage}
\end{table}

\begin{figure*}[t]
\centering
\includegraphics[width=\textwidth]{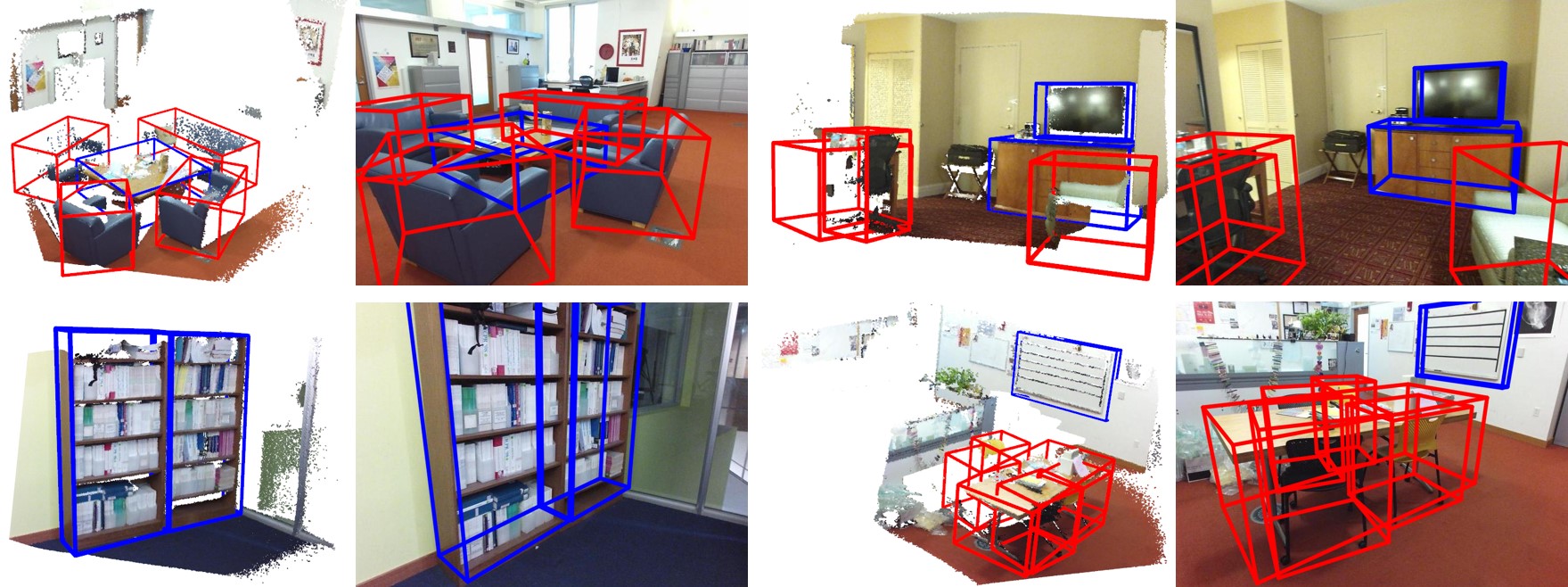}
\caption{\textbf{Visualization of OV-Uni3DETR for open-vocabulary 3D detection} on SUN RGB-D. The red boxes are {\color{red} base} classes and blue boxes are {\color{blue} novel} classes.}
\label{fig:sunrgbdvisual}
\end{figure*}

\begin{figure*}[t]
\centering
\includegraphics[width=\textwidth]{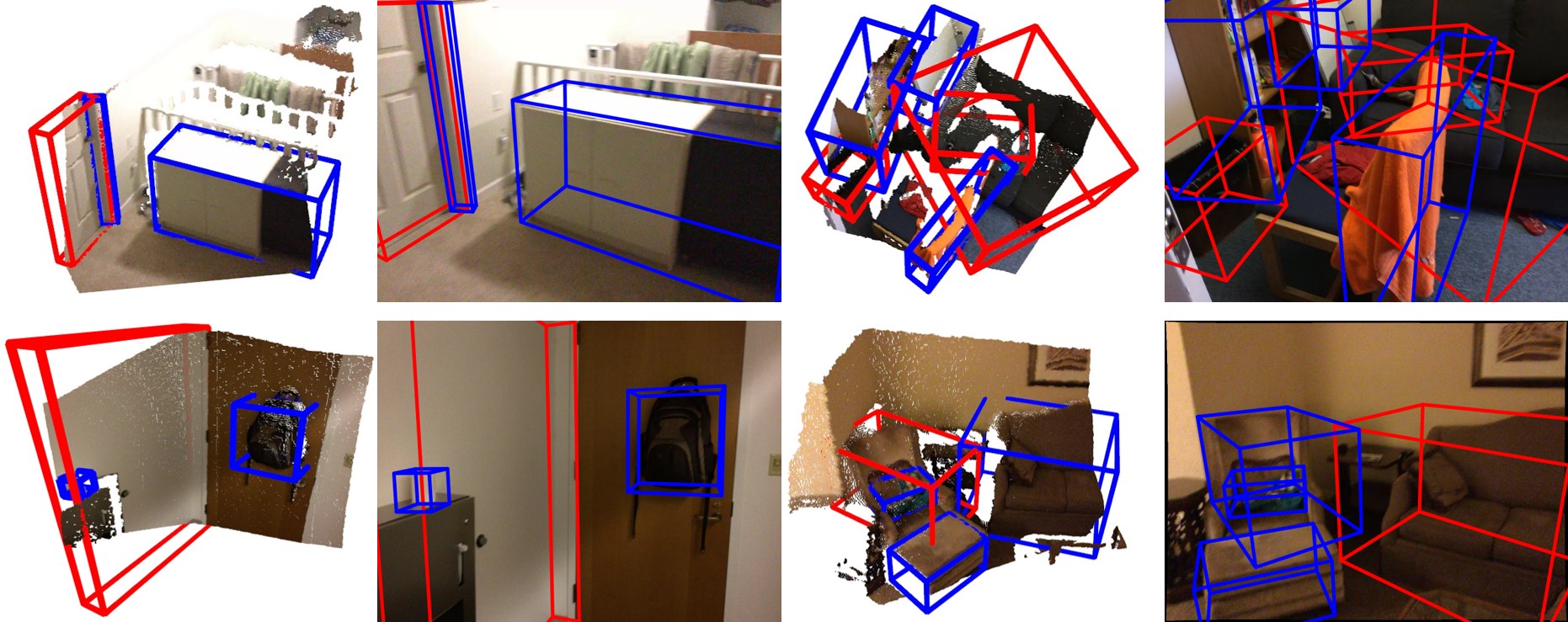}
\caption{Visualization of OV-Uni3DETR for open-vocabulary 3D detection on the \textbf{ScanNet single-view setting}. The point cloud scenes are reconstructed from one single-view depth image. This setting is the same as that from CoDA \cite{cao2023coda}. The red boxes are {\color{red} base} classes and blue boxes are {\color{blue} novel} classes.}
\label{fig:scannetvisual1}
\end{figure*}

\begin{figure*}[t]
\centering
\includegraphics[width=\textwidth]{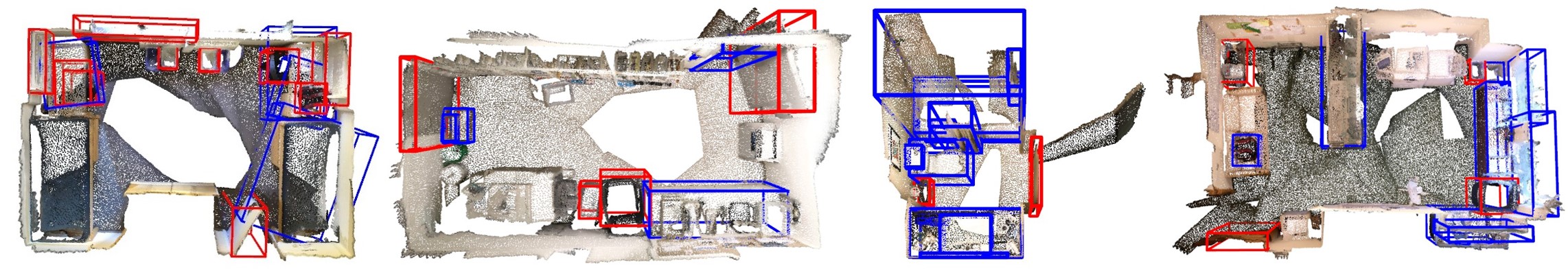}
\caption{Visualization of OV-Uni3DETR for open-vocabulary 3D detection on the \textbf{ScanNet multi-view setting}. Point cloud scenes are reconstructed from plenty of multi-view images. The red boxes are {\color{red} base} classes and blue boxes are {\color{blue} novel} classes.}
\label{fig:scannetvisual2}
\end{figure*}

\begin{figure*}[t]
\centering
\includegraphics[width=\textwidth]{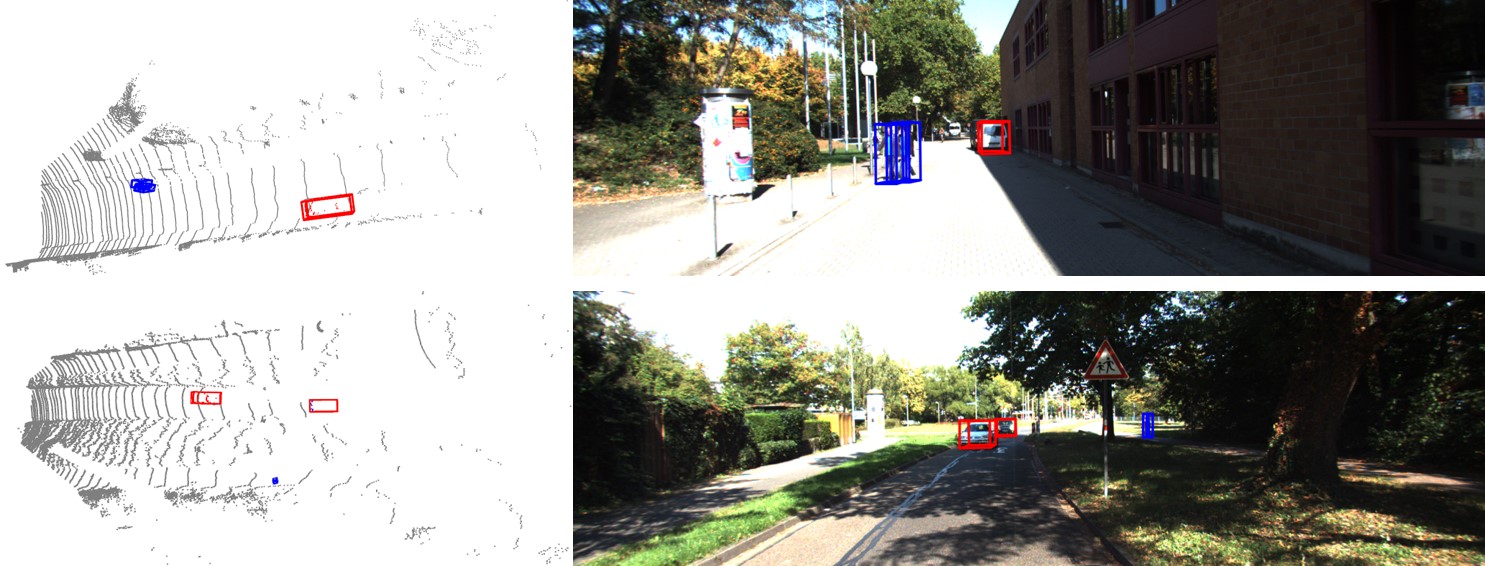}
\caption{\textbf{Visualization of OV-Uni3DETR for open-vocabulary 3D detection} on the KITTI dataset. The red boxes are {\color{red} base} classes and blue boxes are {\color{blue} novel} classes.}
\label{fig:kittivisual}
\end{figure*}

Here we further conduct the ablation study on the 2D detection images we used. We use 2D detection images from the COCO \cite{lin2014microsoft}, ScanNet \cite{dai2017scannet}, Objects365 \cite{shao2019objects365}, OpenImages \cite{kuznetsova2020open}, and LVIS \cite{gupta2019lvis} dataset, together with the corresponding category vocabularies. According to our main paper, 2D detection images benefit RGB-only inference most significantly, so we just conduct experiments with only RGB images here. The 3D detection AP is listed in Tab. \ref{tab:data}.

As we can see, when adopting 2D detection images from the COCO dataset, the performance of 3D open-vocabulary detection is limited, only 3.29\% AP$_{novel}$. The reason is that the vocabulary size of the COCO dataset is still relatively small, only 80 categories. Many of them are outdoor classes, not overlapped with classes in the indoor SUN RGB-D dataset. When we adopt the ScanNet dataset, since it covers the similar indoor scenes, it benefits 3D open-vocabulary detection on the SUN RGB-D dataset more, contributing to the 5.65\% AP$_{novel}$ ultimately. When the vocabulary size of 2D detection images continues to increase, like the OpenImages dataset with 500 classes or the LVIS dataset with 1,230 classes, 3D open-vocabulary detection can be further boosted - 5.89\% AP$_{novel}$ after leveraging images from LVIS. This demonstrates that since 2D detection images contain more images and categories, more abundant information can be utilized after involving them in training. As a result, the performance of 3D open-vocabulary detection, especially in novel class recognition, can be boosted significantly. 

\noindent \textbf{More 2D detection images.} In our experiments on the RUN RGB-D dataset, we utilize about 6,000 2D detection images from the Objects365 dataset. Even if the number of 2D detection images is 6,000, they contain more classes (365 \emph{v.s.} 10 in 3D) and objects (90k \emph{v.s.} 18.5k in 3D). Thus, they are relatively abundant compared to 3D data. We further utilize 12,000 in total. The RGB-only AP$_{novel}$ improves to 6.58\%. In comparison, when 2D detection images are only 6,000, AP$_{novel}$ is 5.41\%. As images are more, AP$_{novel}$ rise turns slower and converges. This demonstrates that the model performance benefits from more abundant 2D detection images.

\subsection{Class-Agnostic 3D Detector}

% \begin{table}[h]
% \centering
% \setlength{\abovecaptionskip}{0pt}
% \setlength{\belowcaptionskip}{0pt}
% \caption{The performance of the class-agnostic 3D detector on the SUN RGB-D dataset .}
% \resizebox{0.4\columnwidth}{!}{
% \begin{tabular}{c|ccc}
% \hline
%  &  AR$_{novel}$ & AR$_{base}$ & AR$_{all}$ \\ 
% \hline
% class-specific &  35.03 & 69.08 & 43.24 \\
% class-agnostic & 55.69 & 88.20 & 68.87 \\
% \hline 
% \end{tabular}
% }
% \label{tab:ca}
% \end{table}

We employ a class-agnostic 3D detector in 2D $\rightarrow$ 3D propagation. We measure its RGB-only recall (AR) in Tab. \ref{tab:ca}. It achieves 20.63\% higher recall compared to the class-specific one (AR$_{novel}$ 35.03\% $\rightarrow$ 55.69\%), because classification is more easily generalized when not specific to particular classes. It can be used in a stand-alone manner in tasks that only require localization but no classification, like obstacle avoidance in robotics. Because of the higher recall metric, the class-agnostic 3D detector can be used in our OV-Uni3DETR to boost class-specific 3D detection.

\section{Visualized Results}

We provide more visualized results on the SUN RGB-D, the ScanNet and the KITTI dataset. OV-Uni3DETR detects novel classes equally well. On the SUN RGB-D dataset (Fig. \ref{fig:sunrgbdvisual}), it recognizes and localizes novel classes like the coffee table in the first example, the tv and cabinet in the second example, the bookshelf and white board in the rest two examples. Besides, it detects seen classes like table, chair successfully. For both point clouds and RGB images, 3D bounding boxes can all be predicted well. On the ScanNet dataset single-view setting (Fig. \ref{fig:scannetvisual1}), our OV-Uni3DETR also detects novel classes like backpack, tissue paper, ottoman well. Meanwhile, for ScanNet multi-view setting (Fig. \ref{fig:scannetvisual2}), where the 3D scene is larger, contained objects are more thus 3D detection is more challenging, our method also detects both base and novel classes in the scene well. On the KITTI dataset (Fig. \ref{fig:kittivisual}), it also detects the novel class pedestrian well. The visualization results further demonstrate the strong ability of OV-Uni3DETR.

\bibliographystyle{splncs04}
\bibliography{main}
\end{document}